\documentclass{article}

\PassOptionsToPackage{numbers, square}{natbib}

\usepackage[final]{neurips_2025}

\usepackage[utf8]{inputenc} 
\usepackage[T1]{fontenc}    
\usepackage{hyperref}       
\usepackage{url}            
\usepackage{booktabs}       
\usepackage{amsfonts}       
\usepackage{nicefrac}       
\usepackage{microtype}      
\usepackage{xcolor}         

\title{\oursimpl{}: Enhancing Agent Programming with Search Over 
Program Execution Paths}

\author{%
  Zhening Li\thanks{
  Work performed as a consultant for Asari AI
  } \\
  Asari AI, MIT CSAIL \\
  \texttt{zhening.li@asari.ai} \\
  \And
  Armando Solar-Lezama \\
  Asari AI, MIT CSAIL \\
  \texttt{asolar@csail.mit.edu} \\
  \AND
  Yisong Yue \\
  Asari AI, Caltech CMS \\
  \texttt{yisong@asari.ai} \\
  \And
  Stephan Zheng \\
  Asari AI \\
  \texttt{stephan@asari.ai} \\
}

\usepackage{graphicx}  
\usepackage{multirow}
\usepackage{listings}
\usepackage{tikz}  
\usepackage{multicol}  
\usepackage{cleveref}  
\usepackage{subcaption}  
\usepackage{bussproofs}  
\usepackage{enumitem}  

\newcommand{\ourslong}{probabilistic angelic nondeterminism}

\newcommand{\ours}{PAN}
\newcommand{\oursimpllong}{\textbf{en}hancing agents with \textbf{comp}iled \textbf{a}gent \textbf{s}earch}
\newcommand{\oursimpl}{\textsc{EnCompass}}

\newcommand{\rev}[1]{#1}

\definecolor{step1color}{HTML}{D43030}
\definecolor{step2color}{HTML}{35982C}
\definecolor{step3color}{HTML}{A136DF}

\definecolor{codegreen}{rgb}{0,0.6,0}
\definecolor{codegray}{rgb}{0.5,0.5,0.5}
\definecolor{codepurple}{rgb}{0.58,0,0.82}
\definecolor{codeblue}{rgb}{0.1,0.2,0.9}
\definecolor{backcolour}{rgb}{0.95,0.95,0.92}

\lstdefinelanguage{AgentPython}{%
  language     = Python,
  emph = {branchpoint, branchpoint_choose, record_score, kill_branch, early_stop_search, searchover, searchover_await, protect, optional_return, record_costs, NoCopy, NeedsCopy},
  morekeywords = {with, as, nonlocal, async, await},
  morestring = [s][\color{orange}]{@encompass}{compile}
}

\lstdefinestyle{mystyle}{
    backgroundcolor=\color{backcolour},   
    commentstyle=\color{codegreen},
    keywordstyle=\color{magenta},
    numberstyle=\tiny\color{codegray},
    stringstyle=\color{codepurple},
    emphstyle=\color{codeblue},
    basicstyle=\ttfamily,
    breakatwhitespace=false,         
    breaklines=true,                 
    captionpos=b,                    
    keepspaces=true,                 
    numbers=left,                    
    numbersep=5pt,                  
    showspaces=false,                
    showstringspaces=false,
    showtabs=false,                  
    tabsize=2,
    columns=fullflexible,
}

\newlist{titledlist}{description}{1}
\setlist[titledlist]{
  style=nextline,   
  labelsep=0pt,     
  leftmargin=0pt,   
  labelindent=0pt,
  itemindent=2em,   
  listparindent=2em,
  font=\bfseries,   
  parsep=0pt,
  itemsep=.7\baselineskip
}

\newenvironment{documentation}%
  {\list{}{\leftmargin=0.25in\rightmargin=0.0in}\item[]}%
  {\endlist}
\newenvironment{ddocumentation}%
  {\list{}{\leftmargin=0.5in\rightmargin=0.0in}\item[]}%
  {\endlist}

\begin{document}

\maketitle

\begin{abstract}
We introduce a new approach to \emph{agent programming}, the development of LLM-based agents.
Current approaches to agent programming often entangle two aspects of agent design: the core workflow logic and the inference-time strategy (e.g., tree search). 
We introduce \textit{\ourslong{}} (\ours{}),
a programming model that disentangles these two concerns,
allowing the programmer to describe the agent workflow and
independently experiment with different inference-time strategies
by simply changing a few inputs.
We provide an implementation of \ours{} in Python as the \oursimpl{} framework, which uses a Python decorator to compile agent workflow programs into a search space.
We present three case studies that demonstrate how the framework lets the programmer quickly improve the reliability of an agent and easily switch between different inference-time strategies, all with little additional coding.

\end{abstract}

\section{Introduction}

Recent work has shown the power of scaling inference-time compute for LLMs
\citep{snell2025scaling,welleck2024metageneration},
where popular strategies include best-of-$N$ sampling \citep{cobbe2021bon,brown2024bon,li2022alphacode},
refinement \citep{madaan2024selfrefine,shinn2024reflexion},
and tree search \citep{yao2024tot,xie2024beam}.
In LLM-based agents ---
systems that define how LLMs and other components interact to solve a task ---
these same strategies have become common ways of improving performance and reliability.
Furthermore, several works have demonstrated the
utility of applying sophisticated search and backtracking strategies
in AI agents to improve performance in various tasks
\citep{yao2024tot,zhou2023lats,koh2024agenttreesearch,antoniades2025swesearch,yang2024leandojo,yamada2025aiscientistv2}.

While various frameworks
have been developed to simplify the low-level interaction between the program and the LLM
\citep{chase2025langchain,khattab2024dspy,beurer2023lmql,wu2024autogen},
a framework for agent inference-time strategies has been absent.
Our goal is to develop an \textit{inference-time strategy framework}:
a framework that makes it easy to experiment with different inference-time strategies
independently of the design and implementation of the underlying agent workflow.
Such a framework is intended not to replace,
but to be used in conjunction with
LLM prompting and tool use frameworks,
such as LangChain \citep{chase2025langchain} or DSPy \citep{khattab2024dspy}.



We target ``program-in-control'' style agents,
where one defines the workflow in code and uses the LLM to accomplish specific subtasks \citep{yamada2025aiscientistv2,wang2023hypothesis,ibrahimzada2025alphatrans,zhou2024selfdiscover,lu2024aiscientist}.\footnote{
This ``program-in-control'' style contrasts with the ``LLM-in-control'' style where
the LLM
decides the full sequence of operations (tool calls)
in the workflow
\citep{yao2024tot,xie2024beam,zhou2023lats,koh2024agenttreesearch,antoniades2025swesearch,yang2024leandojo}.}
In these agents, inference-time strategies have traditionally
been limited to sampling and refinement loops 
\citep{madaan2024selfrefine,shinn2024reflexion,wang2023hypothesis,ibrahimzada2025alphatrans},
whereas more sophisticated strategies such as beam search and tree search
have been rarely explored \citep{yamada2025aiscientistv2}.
We identify the key bottleneck to be the entanglement of 
the inference-time scaling strategy with the core workflow logic
when programming the agent.
Programmers typically bake the inference-time strategy into the agent workflow
\citep{yamada2025aiscientistv2,wang2023hypothesis,ibrahimzada2025alphatrans,lu2024aiscientist},
which is inflexible, reduces readability,
and limits the kinds of inference-time strategies
that can be easily implemented.
Therefore, we aim to design a framework that
cleanly separates the representation of the core workflow logic
from the inference-time scaling strategy.
The programmer could then make minimal modifications to their agent to
flexibly experiment with different inference-time strategies. Also,
different agents would no longer require
custom implementations of the same inference-time strategy,
but can instead reuse a common implementation.

\begin{figure*}[tb]
\begin{center}
\centerline{\includegraphics[width=\textwidth]{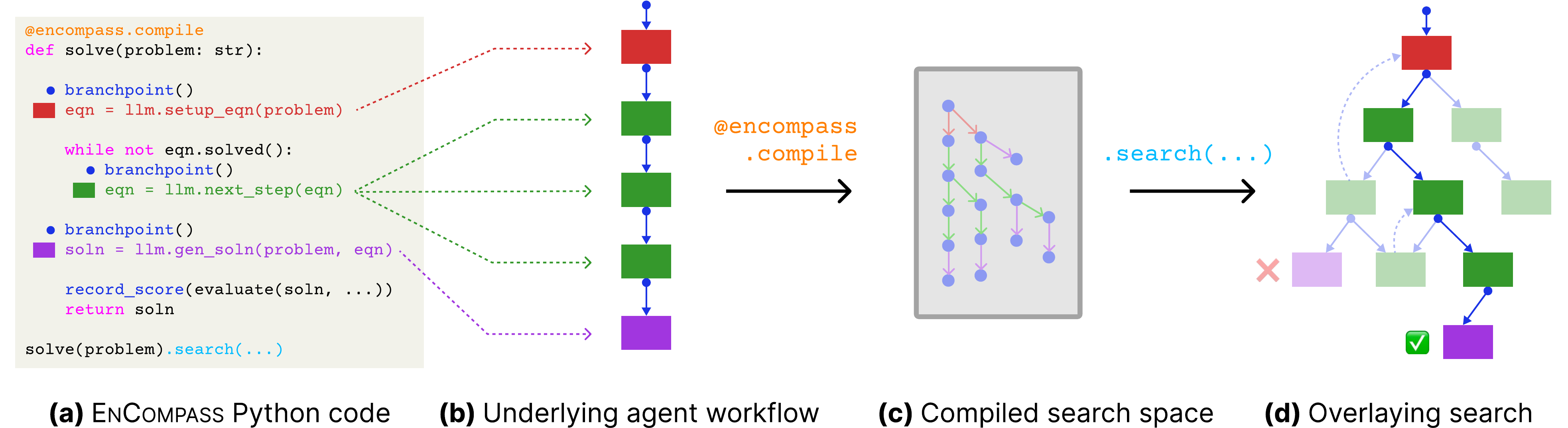}}
\caption{An \oursimpl{} program specifies an agent workflow,
which is compiled into a search space object, and inference-time scaling
is accomplished through search over the nondeterministic execution paths
of the agent workflow.}
\label{fig:ours}
\end{center}
\vskip -0.25in
\end{figure*}

Our key insight is that inference-time strategies can be viewed as instances of
\textit{search over different execution paths of a nondeterministic program}.
We developed the \oursimpl{} Python programming framework (``\oursimpllong{}''), depicted in \Cref{fig:ours}.
\Cref{fig:ours}a and \Cref{fig:ours}b show an agent program and its corresponding workflow, respectively.
The user specifies the ``locations of unreliability'' in their agent source code
using \lstinline|branchpoint()| statements.
A location of unreliability is an operation such as an LLM call
where repeated invocations produce outputs
of varying quality.
Since these different outputs give rise to multiple possible
futures of the program's execution,
the program has a tree of possible execution paths.
\oursimpl{} compiles the program into a search space object (\Cref{fig:ours}c)
so that search can be conducted over this tree of execution paths
to find the path with the highest score (\Cref{fig:ours}d).
We call this programming model \textit{\ourslong{}} (\ours{}).
As a form of angelic nondeterminism~\cite{floyd1967nondeterministic},
\ours{} lets the programmer write their program pretending
the unreliable operations always produce good outputs,
and the runtime searches the space of
possible execution paths for one where the operations
indeed produced good outputs.

Our work makes the following concrete contributions: 
\begin{itemize}
    \item We introduce the \ours{} programming model (\Cref{sec:ours}),
    which uses angelic nondeterminism to separate inference-time algorithms
    (search policy) from the underlying logic of the agent
    (specification of the search space).
    \item We present \oursimpl{}, a Python library that implements \ours{} (\Cref{sec:oursimpl}), providing 1.\ primitives like \lstinline|branchpoint()| that the programmer can use inside their \oursimpl{} function,
    2.\ a Python function decorator that compiles an \oursimpl{} function into a search space
    object at run-time, and 3.\ common search algorithms, as well as an interface for implementing custom search algorithms.
    \item We illustrate how \oursimpl{} provides a unifying framework for common inference-time strategies and agentic patterns, which are special cases of search over nondeterministic execution paths of \oursimpl{} programs (\Cref{sec:agent_patterns}). 
    \oursimpl{} also provides a natural generalization of these inference-time strategies.
    \item We present case studies showing how \oursimpl{} enables
    easy experimentation of various inference-time search strategies
    over an underlying agent workflow, allowing one to quickly
    identify the best-performing strategy (\Cref{sec:case_studies}).
    \oursimpl{}
    opens up new possibilities for inference-time scaling
    of program-in-control style agents,
    where
    inference-time strategies that were previously
    considered too cumbersome to implement are
    now made possible by \oursimpl{}.
\end{itemize}



\section{\oursimpl{}, a Python framework for \ours} \label{sec:ours_oursimpl}

In this section, we introduce the \ours{} programming model (\Cref{sec:ours}) and
describe its Python implementation in the \oursimpl{} framework (\Cref{sec:oursimpl}).
For simplicity, we will ignore the feature of memory sharing across different program execution paths
(see \Cref{sec:refinement}).
The documentation for \oursimpl{} is in \Cref{app:documentation} 
and the \oursimpl{} compiler is described in \Cref{app:compiler_details}.

\subsection{Probabilistic angelic nondeterminism (\ours{})} \label{sec:ours}

The core idea of \ours{} is to search over the tree of possible execution paths
of a probabilistic program --- where some operations (e.g., LLM calls) have randomness --- to find the path that optimizes a user-specified objective.
Given a probabilistic program with branchpoints at certain locations in the program,
we model its computation as
a Markov chain over the space of possible program states.
The Markov chain consists of the following components:
\begin{itemize} 
    \item Branchpoints and the end of the program
    constitute a set of \textit{marked locations} in the program.
    In \Cref{fig:ours}, branchpoints are denoted by blue dots
    \,\raisebox{0.3ex}{\tikz{\fill[codeblue] (0,0) circle (1.5pt);}}\,.
    \item
    A \textit{program state} is a pair consisting of
    a marked location of the program and a \textit{memory state},
    which is a mapping from variables to values.
    \item
    The code that executes from one marked location to the next
    defines a \textit{probabilistic transition function}
    that maps from a program state to the next program state.
    Program states at the end of the program are \textit{final},
    i.e., they have no next states.
    In \Cref{fig:ours},
    transitions are denoted by colored boxes
    \,\raisebox{0.1ex}{\tikz \fill[step1color] (0,0) rectangle (9pt,6pt);}
    \raisebox{0.1ex}{\tikz \fill[step2color] (0,0) rectangle (9pt,6pt);}
    \raisebox{0.1ex}{\tikz \fill[step3color] (0,0) rectangle (9pt,6pt);}\,.
    \item The \textit{initial state} is the program state resulting from executing the program
    from the start until hitting the first branchpoint.
\end{itemize}

Normally, executing the probabilistic program results in one sampled trajectory of program states
(\Cref{fig:ours}b).
In \ours{}, however, we \textit{search} over the space of possible trajectories
(\Cref{fig:ours}d).
Our search tree initially has just one node: the initial program state. 
At every step, the search policy chooses a node 
in the current search tree, makes a copy of the program state
stored at that node,
samples a next program state according to the probabilistic transition function, 
and adds it to the search tree as a child of that node.
The goal is to reach a final program state 
that optimizes a user-specified objective.

Note that search here is formulated differently from the usual graph search formulation
because we don't have access to all the children of any given node --- we can only stochastically \textit{sample}
children of a parent node. 
However, existing graph search algorithms can be converted to algorithms in \ours{}
by specifying each node's \textit{branching factor}, i.e., the number of children to sample.
For example, depth-first search (DFS) with branching factor 3 involves
sampling 3 next states from the current state and recursing on each child.

This way of adapting graph search algorithms is currently the dominant approach
in LLM-based agents that have tree search with an unenumerable action space of LLM outputs
\citep{yao2024tot,yang2024leandojo,zhang2024matholymcts,wan2024llmmcts}.
However, we believe it is worth exploring search strategies beyond
fixing the branching factor in an existing graph search algorithm,
and Case Study 3 (\Cref{sec:reflexion}) explores this
direction by showing that a simple strategy --- 
repeatedly choosing the highest-scoring program state
and sampling one next state --- can work quite well.

\subsection{\oursimpl{}} 
\label{sec:oursimpl}

The \oursimpl{} framework provides an instantiation of the \ours{} programming model in Python.
It is implemented as the \lstinline|@encompass.compile| function decorator,
which makes several new primitive keywords available in the body of the decorated function; the
full list is given in \Cref{app:primitives}.
The decorator compiles the function body into a search space object, which provides an interface for implementing search algorithms (\Cref{app:compiled,app:checkpoint}).
The compiler is described in \Cref{app:compiler_details}.

\paragraph{Core primitives} The two most important primitives that
are available in the body of an \oursimpl{}-decorated function
are \lstinline|branchpoint()| and \lstinline|record_score()|.

\noindent\lstinline[emph={[2]branchpoint_params}, emphstyle={[2]\itshape}]|branchpoint(**branchpoint_params)|

\begin{documentation}
This statement marks a \ours{} branchpoint (\Cref{sec:ours}),
a location in the program where the program state is added as a new node in the search tree
and the program's execution may branch into multiple execution paths.

Branchpoint parameters provide information to the external search algorithm about the branchpoint.
For example,
\lstinline|branchpoint(name="foo")|
gives the branchpoint a name that can be used to refer
to the branchpoint in the search algorithm.
\end{documentation}

\noindent\lstinline[emph={[2]score}, emphstyle={[2]\itshape}]|record_score(score)|

\begin{documentation}
This records the numerical ``score'' used to guide the search process
in many search algorithms (e.g., the heuristic in best-first search and value function in MCTS).
Furthermore, the final score (the last score recorded before returning)
usually specifies the final evaluation score to be maximized by the search algorithm.
\end{documentation}

\vspace{0.5em}

\paragraph{Inference-time search}
Having defined an \oursimpl{}-decorated function \lstinline[emph={[2]func}, emphstyle={[2]\itshape}]|func|,
the programmer can now apply search over its nondeterministic execution paths by calling

\noindent\lstinline[emph={[2]func, algo, search_config}, emphstyle={[2]\itshape}]|func(...).search(algo, **search_config)|

where \lstinline[emph={[2]algo}, emphstyle={[2]\itshape}]|algo| is a string such as \lstinline|"dfs"| or \lstinline|"beam"|
specifying the search algorithm.
This returns the function's return value on
the best execution path that search algorithm \lstinline[emph={[2]algo}, emphstyle={[2]\itshape}]|algo| could find.
\Cref{app:search} lists all algorithms that \oursimpl{} provides
out-of-the-box.

\paragraph{Custom search algorithms}
The user can also define and register their custom search algorithm
so that it can be invoked through the same search interface.
The \lstinline|Checkpoint| class wraps the program state
and provides an interface for implementing custom search algorithms.
Its \lstinline|step()| method samples a next program state: it
resumes execution of the program from the current state
until hitting the next branchpoint or a return statement,
returning the new program state (cf.\ the probabilistic transition function from \Cref{sec:ours}).
The \lstinline|Checkpoint| object's \lstinline|score| attribute
contains the score of the program state as recorded through \lstinline|record_score()|.
See \Cref{app:checkpoint} for more details.

\section{Agent inference-time strategies in \oursimpl{}} \label{sec:agent_patterns}

While \oursimpl{} appears most suitable for implementing
tree search in agents,
other common inference-time strategies
can also be cleanly implemented
as search in \oursimpl{}.
Furthermore, natural generalizations of these strategies that are otherwise
difficult to implement are also easily represented in \oursimpl{}.

\subsection{Best-of-$N$ sampling and beam search} \label{sec:bestofn}

Given an agent \verb|agent_forward(...)| and an evaluator \verb|evaluate(...)|
that evaluates the output of the agent, \textit{best-of-$N$} (BoN) samples $N$ times and chooses
the output with the highest evaluation score.
%
In \oursimpl{}, this is done by adding a branchpoint at the beginning
of the function and recording the evaluation score at the end:
\begin{lstlisting}[basicstyle=\ttfamily\scriptsize]
@encompass.compile
def agent_forward(...):
    branchpoint()
    ...  # Original body of agent
    record_score(evaluate(result))
    return result

result = agent_forward(...).search(...)
\end{lstlisting}

This defines a search tree with depth 1, where almost any search algorithm
would sample several children from the root node and return the best child,
thus reproducing best-of-$N$ sampling.

We call the above \textit{global best-of-$N$} (GBoN) to contrast it with
\textit{local best-of-$N$} (LBoN),
where an agent with multiple verifiable steps has best-of-$N$ sampling applied to each of them.
In \oursimpl{}, this is implemented by adding \lstinline|branchpoint()| before each step
and applying beam search with beam width 1:
\begin{lstlisting}[basicstyle=\ttfamily\scriptsize, ]
@encompass.compile
def agent_forward(...):
    branchpoint()
    ...  # Step 1
    record_score(evaluate_step1(...))
    branchpoint()
    ...  # Step 2
    record_score(evaluate_step2(...))
    ...
    branchpoint()
    ...  # Step k
    record_score(evaluate_stepk(...))
    return stepk_result

N = ...  # the "N" in best-of-N
result = agent_forward(...).search("beam", beam_width=1, default_branching=N)
\end{lstlisting}
Note that the two types of best-of-$N$ sampling described in this section---\textit{global}
and \textit{local} sampling---are the two limiting cases of beam search.
Global best-of-$N$ sampling is beam search with beam width $N$
and branching factor 1,\footnote{
Except that the root node has branching factor $N$.}
whereas local best-of-$N$ sampling is beam search with beam width 1 and branching factor $N$.
General beam search can thus be viewed as interpolating between global and local resampling.
This has the benefit of effectively constraining the search space with local verification
while also not losing global variety.
Increasing the branching factor makes sure each step is completed correctly
to help prevent compounding errors, while
increasing the beam width can help increase variety and thus improve reliability
to mitigate potential errors made in earlier steps.
In Case Study 1 (\Cref{sec:codetrans}), we empirically demonstrate that
beam search indeed scales better than global best-of-$N$ or local best-of-$N$ alone
in complex agent workflows.

The \oursimpl{} implementation of beam search over an agent workflow
also benefits from flexibility in modifying the step granularity.
Increasing the granularity (dividing steps up into smaller substeps)
or decreasing the granularity (merging multiple steps into one)
is as simple as adding or removing branchpoints in \oursimpl{},
whereas a plain Python implementation would require structural changes to the code.


\subsection{Refinement and backtracking with memory} \label{sec:refinement}

Refinement can be viewed as sampling but with additional feedback from past sampling attempts.
In \oursimpl{}, this is accomplished by adding a branchpoint to generate multiple samples
and a memory of past attempts shared across the different sampled execution paths:
\begin{lstlisting}[basicstyle=\ttfamily\scriptsize]
@encompass.compile
def agent_forward(...):
    ... # stuff that comes before
    # Step with refinement
    feedbacks: NoCopy = []
    branchpoint()
    result = do_step(..., feedbacks)
    score, feedback = get_score_and_feedback(result)
    feedbacks.append(feedback)
    record_score(score)
    ... # stuff that comes after

result = agent_forward(...).search("beam", beam_width=1, default_branching=n_refine + 1)
\end{lstlisting}
Here, the \lstinline|NoCopy| type annotation tells the \oursimpl{} compiler
that 
the different execution paths should share the same reference
to the \lstinline|feedbacks| variable, so that appending \lstinline|feedback|
is seen across all branches.\footnote{
This effect is lost if \lstinline|feedbacks.append(feedback)| is replaced with
\lstinline|feedbacks = feedbacks + [feedback]|, since that creates a new list
instead of modifying the original one.
}

By adding another \lstinline|branchpoint()| right before ``\lstinline|feedbacks: NoCopy = []|'',
we create multiple parallel refinement loops, thus interpolating between
fresh sampling and refinement and maintaining variety that may otherwise be lost
from an agent that focuses too heavily on the past feedback.
This is not unlike how beam search interpolates between global best-of-$N$
and local best-of-$N$ (\Cref{sec:bestofn}).
In Case Study 3 (\Cref{sec:reflexion}), we demonstrate how a different approach
to interpolating between refinement and sampling --- by adding branchpoints to a
refinement loop written in plain Python --- can result in better scaling than refinement alone.

Note that refinement is the simplest case of \textit{backtracking with memory}:
backtracking to a previous step while remembering what happened in previous attempts.
In \oursimpl{}, the general pattern for backtracking with memory is to create
a shared mutable data structure right before a branchpoint,
which serves as a memory shared across all execution paths that follow.

\subsection{Self-consistency and group evaluation} \label{sec:self-consistency}

Given an agent program \verb|agent_forward(input)|,
\textit{self-consistency} samples $N$ times and chooses
the output that appeared the most times (the majority vote) \citep{wang2023selfconsistency}.
This can be implemented as best-of-$N$ sampling with an evaluation function
that evaluates a group of results at once.
The \oursimpl{} \lstinline|record_score()| supports this:
\begin{lstlisting}[basicstyle=\ttfamily\scriptsize, multicols=2]
def majority_vote(results):
    counts = defaultdict(int)
    for result in results:
        counts[result] += 1
    return [
        counts[result] for result in results
    ]

@encompass.compile
def agent_forward(...):
    branchpoint()
    result = ...
    record_score(majority_vote, result, label=None)
    return result

result = agent_forward(...).search(...)
\end{lstlisting}

In general, allowing the evaluation function to evaluate a group of results at once
is helpful when it is difficult to evaluate one result on its own.
Another example of this is CodeT \citep{chen2023codet},
which evaluates a group of LLM-generated code samples against multiple LLM-generated unit test cases
by considering both the number of unit test pass rate and agreement among code samples
on which test cases they pass.

Inference-time strategies like self-consistency and CodeT
are examples of the more general \textit{search with evaluation of a group of execution paths in tandem}.
When one writes

\noindent\lstinline[emph={[2]group_evaluator, evaluation_target, group_label}, emphstyle={[2]\itshape}]|record_score(group_evaluator, evaluation_target, label=group_label)|

the scores of all program states where \lstinline|record_score()|
was called with label
\lstinline[emph={[2]group_label}, emphstyle={[2]\itshape}]|group_label|
are computed as \lstinline[emph={[2]group_evaluator, evaluation_targets}, emphstyle={[2]\itshape}]|group_evaluator(evaluation_targets)|,
where \lstinline[emph={[2]evaluation_targets}, emphstyle={[2]\itshape}]|evaluation_targets| is the list of the \lstinline[emph={[2]evaluation_target}, emphstyle={[2]\itshape}]|evaluation_target|
variables across all the program states.

\section{Case studies} \label{sec:case_studies}

We implemented and extended 3 program-in-control style agents
from the literature in \oursimpl{}. These case studies aim to answer
the following research questions:
\begin{itemize}
    \item Does \oursimpl{} make it easier to implement
    inference-time strategies and search in program-in-control style agents, and if so, how?
    \item Does \oursimpl{} simplify experimenting
    with different inference-time strategies and search in program-in-control style agents, and if so, how?
\end{itemize}
Our case studies suggest that \oursimpl{}
enables the exploration of inference-time strategies
that are otherwise left unexplored due to
their complexity of implementation
--- potentially unlocking better scaling laws.

Case Study 1 is our main case study and is presented in the main text here.
Case Studies 2 and 3 are smaller and more didactic in purpose,
and are presented in \Cref{app:case_studies}.

In \textbf{Case Study 1} (\Cref{sec:codetrans}), 
we implement a Java-to-Python code repository translation agent
with a high-level architecture based on that of Syzygy \citep{shetty2024syzygy}.
We then add branchpoints before LLM calls and, \rev{by toggling a few parameters,}
we experiment with a variety of search strategies
including local/global best-of-$N$ sampling and beam search
at the file level and individual method level.
\rev{We demonstrate these experiments on Java repositories from
the MIT OCW Software Construction class.}
We find that beam search outperforms simpler sampling strategies,
thus demonstrating how one can use \oursimpl{} to discover better inference-time scaling laws.
Furthermore, we show how the equivalent plain Python implementation
of the \oursimpl{} agent involves defining the search graph as a state machine,
where the agent workflow is significantly obscured and modularity is compromised,
whereas \oursimpl{} solves these issues.

In \textbf{Case Study 2} (\Cref{sec:hypo_search}), \rev{we implement a simplified Hypothesis Search agent \citep{wang2023hypothesis}.}
We start with a simple agent with two LLM calls.
By adding a branchpoint before each LLM call and applying
multithreaded BFS out of the box,
we reproduce a parallelized version of Hypothesis Search.
We demonstrate how to use \oursimpl{} to experiment with different search strategies (BFS vs.\ global best-of-$N$),
and find that they perform equally well on a subset of the ARC benchmark \citep{chollet2019arc},
the benchmark that Hypothesis Search used.
We show how, despite the simplicity of the original agent,
the equivalent program in plain Python already noticeably
obscures the underlying agent workflow.

In \textbf{Case Study 3} (\Cref{sec:reflexion}), we start with Reflexion \citep{shinn2024reflexion},
a simple agent with a refinement loop.
We add a branchpoint at the beginning of the agent and
at the beginning of the body of the refinement loop,
and apply both global best-of-$N$ and a variant of best-first search.
Following the original Reflexion paper,
we evaluate on LeetCodeHard.
We find that increasing $N$ in best-of-$N$
or the number of search steps in best-first search
scales better than increasing the number of refinement iterations in vanilla Reflexion.
We also show how the equivalent program in plain Python obscures the
control flow and data flow of the underlying agent.

\addtolength{\tabcolsep}{-0.225em} 
\begin{table}[tb]
\small
\centering
\caption{Code modifications to implement search in our case studies,
without \oursimpl{} vs.\ with \oursimpl{}.
Metrics include the number lines/words added, changed\textsuperscript{\textit{a}}, and removed,
the number of new function definitions, and the number of lines of the original code
where the indentation level was changed.
For context, we also give the number of lines of code used to implement the core logic\textsuperscript{\textit{b}}
of the original base agent.
All code is found in \Cref{app:plain-python} with the modifications annotated.
\\
{\footnotesize
\textsuperscript{\textit{a}} This excludes changes to the indentation level of existing code.
\textsuperscript{\textit{b}} The ``core logic'' is defined as
the functions that require modification when implementing search,
hence excluding unmodified code like helper/utility functions and prompt templates.
}
}
\vspace{0.05in}
\label{tab:code_changes}
\begin{tabular}{llcccccc}
\toprule
\multirow{2}{*}{\textbf{Case Study}} & & \textbf{Added} & \textbf{Changed} & \textbf{Removed} & \textbf{New} & \textbf{Indent} \\
&  & lines (words) & lines (words) & lines (words) & \textbf{f'ns} & \textbf{changed} \\
\midrule
1. Code Repo Translation & $-$\oursimpl & +423 (+2735) & 24 (-62/+186) & -9 (-28) & +20 & 189 \\
\quad LoC = 597         & $+$\oursimpl    & \textbf{+75 (+514)}   & \textbf{8 (-0/+40)}    & \textbf{-0 (-0)}  & \textbf{+1} & \textbf{0}  \\
\midrule
2. Hypothesis Search & $-$\oursimpl & +21 (+120) & 3 (-1/+13) & -0 (-0) & +2 & 10 \\
\quad LoC = 11  & $+$\oursimpl    & \textbf{+8 (+27)}   & \textbf{1 (-0/+9)}  & -0 (-0) & \textbf{+0} & \textbf{0}  \\
\midrule
3. Reflexion & $-$\oursimpl & +27 (+181) & 6 (-13/+31) & -0 (-0) & +2 & 8 \\
\quad LoC = 20  & $+$\oursimpl    & \textbf{+9 (+32)}   & \textbf{3 (-4/+13)}  & -0 (-0) & \textbf{+0} & \textbf{0}  \\
\bottomrule
\end{tabular}
\vskip -0.1in
\end{table}
\addtolength{\tabcolsep}{0.225em} 

\Cref{tab:code_changes} and \Cref{app:plain-python}
compare the code modifications required to implement search
with \oursimpl{} vs.\ without \oursimpl{}.
On average, \oursimpl{} saves 3--6x of coding in terms of the number of lines/words that are added or changed.

\rev{Note that since \oursimpl{} targets program-in-control style
agents, our case studies do not include benchmarks of LLM-in-control style
agents such as SWEBench \citep{jimenez2024swebench} or WebArena \citep{zhou2023webarena}.
}

\subsection{Case Study 1: Code Repository Translation Agent} \label{sec:codetrans}

In this case study, we demonstrate how to use \oursimpl{}
to add branchpoints and implement search in a Java-to-Python code repository translation agent
based on the Syzygy agent architecture \citep{shetty2024syzygy}.
By comparing with the equivalent plain Python implementation,
we identify several concrete benefits of the separation of concerns offered by \oursimpl{}.
We also demonstrate experimenting with different search strategies on
one repository to find the best-performing strategy (``fine-grained'' beam search),
and we apply this strategy to other repositories to obtain
strong performance compared to simpler strategies (global/local best-of-$N$).

\paragraph{Base agent}
We built an agent that translates a Java repository into Python (\Cref{lst:code_trans_base}).
The agent translates the repository file-by-file in dependency order.
For each file, the agent calls the LLM to write the skeleton of the Python file,
and for each Java method the agent calls the LLM to translate it into Python.
Every translation is followed by validation of the translation
by 1) asking the LLM to write a script that generates random test case inputs;
2) asking the LLM to write Java code to run the Java method on those inputs;
3) asking the LLM to write Python code to run the translated Python method on those inputs;
and 4) comparing the Python and Java outputs to see if they match.

\paragraph{The \oursimpl{} agent}
In \oursimpl{}, we modify the base agent by adding a branchpoint before each of the
5 LLM calls present in the program (\Cref{lst:code_trans_ours}).
To prevent different branches of the search from overwriting the same folder,
we use Git to manage the repository, and write a wrapper
\lstinline|branchpoint_git_commit()|
around the built-in \lstinline|branchpoint()|
(\Cref{lst:code_trans_ours}, L5--15).
We consider search at two different levels of the translation workflow:
the file level (``coarse''), and the method level (``fine'').
By adjusting the search parameters, we experimented with different search strategies at each level
as well as different parameters to the search strategies.
We applied 6 combinations of search strategies: ``global best-of-$N$'',
``local best-of-$N$ (coarse)'', ``local best-of-$N$ (fine)'',
``beam (coarse)'', ``local best-of-$N$ (coarse) + beam (fine)'', and ``beam (coarse) + beam (fine)''.

This was as simple as changing a couple of parameters:
the file-level search strategy is specified at line 278 of \Cref{lst:code_trans_ours}
and the method-level search strategy is specified at line 264.
\Cref{sec:bestofn} explains the search algorithm and parameters
passed to the \lstinline|.search_multiple(...)| method
to implement global BoN, local BoN, and beam search.

\paragraph{Comparison with equivalent plain Python}
We demonstrated how \oursimpl{} lets an agent programmer easily switch between different search algorithms.
To replicate this flexibility in plain Python, we need to explicitly define the search graph
that the \oursimpl{} function defines.
The search graph takes the form of a state machine where the states correspond
to the branchpoints and the transitions follow the control flow of the program.
We maintain a dictionary \lstinline|frame| with all the local variables of the program as we go through
the transitions of the state machine.
The result is \Cref{lst:code_trans_plain}, which is long and difficult to read,
so we illustrate this with a simplified version of our code repository translation agent
that iterates through functions in a source file, translating each of them one-by-one:
\begin{lstlisting}[basicstyle=\ttfamily\scriptsize]
@encompass.compile
def translate_functions(source):
    for source_fn in source:
        branchpoint()
        target_fn = translate(source_fn)
        compile_success = compile_(target_fn)
        record_score(compile_success)

        branchpoint()
        unit_test_score = run_unit_test(target_func)
        record_score(unit_test_score)
\end{lstlisting}
The equivalent state machine in plain Python is given here
with minor simplifications.
\begin{lstlisting}[basicstyle=\ttfamily\scriptsize]
class State(Enum):
    TRANSLATE = auto()
    UNIT_TEST = auto()

def step(state: State, frame: dict[str, Any]):
    frame = frame.copy()

    if state == State.TRANSLATE:
        frame["target_fn"] = translate(frame["source_fn"])
        compile_success = compile_(frame["target_fn"])
        return State.UNIT_TEST, frame, compile_success

    if state == State.UNIT_TEST:
        unit_test_score = run_unit_test(frame["target_fn"])
        frame["source_fn"] = next(frame["source"])
        return State.TRANSLATE, frame, unit_test_score
\end{lstlisting}
Notice that the high-level control flow of ``repeatedly translate and unit-test the translation''
is no longer obvious from the code;
it is difficult to know whether any given variable access \lstinline|frame[...]|
might throw a \lstinline|KeyError|;
and linters and static type checkers can't be applied because variables are accessed through
the \lstinline|frame| dictionary.
Furthermore, simple changes to the \oursimpl{} function such as moving or removing
a branchpoint would require significant structural changes to the state machine code
that further create an opportunity for bugs.
All these issues are exacerbated as we increase the complexity of the agent program, so the state machine approach to defining agent search graphs is not scalable.
This can be seen in \Cref{lst:code_trans_plain}
(\Cref{app:python_case3}), which
applies the state machine approach to the original code repository translation agent.

\paragraph{Evaluation setup}
To make it affordable to run comprehensive experiments comparing the scaling behaviors
of various inference-time strategies, we first validated on a small repository
consisting of 622 lines of Java code.
The repository contains solutions to the first homework (\texttt{ps0})
from the Spring 2016 version of the MIT Software Construction class available on MIT OpenCourseWare
\citep{savenkov2017mit6005,miller2016mit6005}.

Because of the scarcity of test cases in the original repository,
we use \textit{self-validation (\%)} as the evaluation metric,
which is calculated as the percentage match of the Python and Java outputs
on the automatically generated test inputs, averaged across all translated non-test methods.
If any step of the validation process failed (e.g., test input generation),
then the match percentage is considered to be $0$.

After identifying the inference-time strategy that scales best on \texttt{ps0},
we evaluated it on the other 4 repositories from the class (\texttt{ps1} to \texttt{ps4}).
Each of them contains between 1100 and 1900 lines of code,
and all 4 repositories combined contain 5756 lines of code.

For all experiments, we set the LLM temperature to $0.0$
for the base agent (no inference-time strategies),
and $0.5$ for the \oursimpl{} agent (with inference-time strategies).

\begin{figure}[t!]
\vskip -0.2in
\begin{center}
    \begin{subfigure}[b]{0.45\textwidth}
    \centering
    \centerline{\includegraphics[width=\textwidth, trim={0.25cm 0.5cm 0.5cm 0.5cm}, clip]{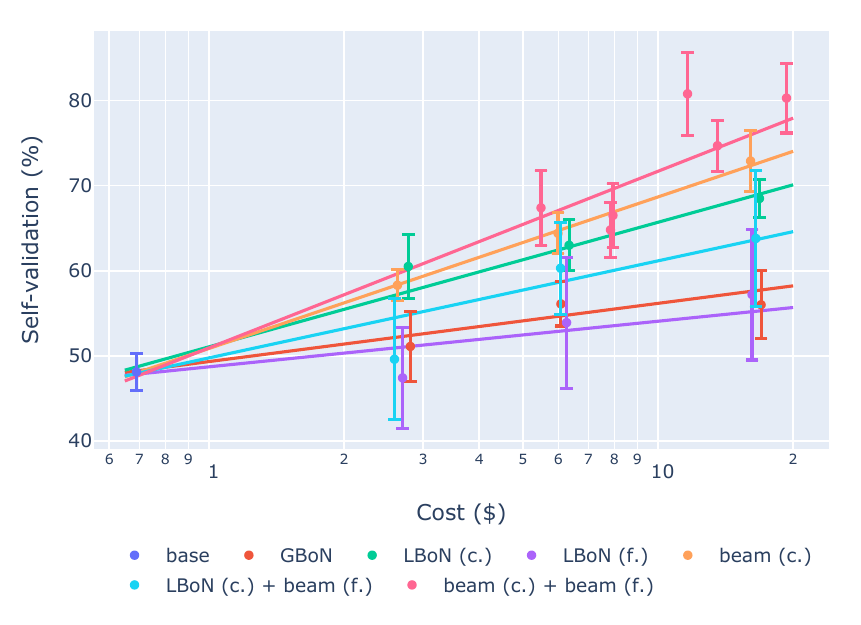}}
    \caption{}
    \label{fig:code-trans-ps0}
    \end{subfigure}
    \hfill
    \begin{subfigure}[b]{0.45\textwidth}
    \centering
    \centerline{\includegraphics[width=\textwidth, trim={0.25 0.5cm 0.5cm 0.5cm}, clip]{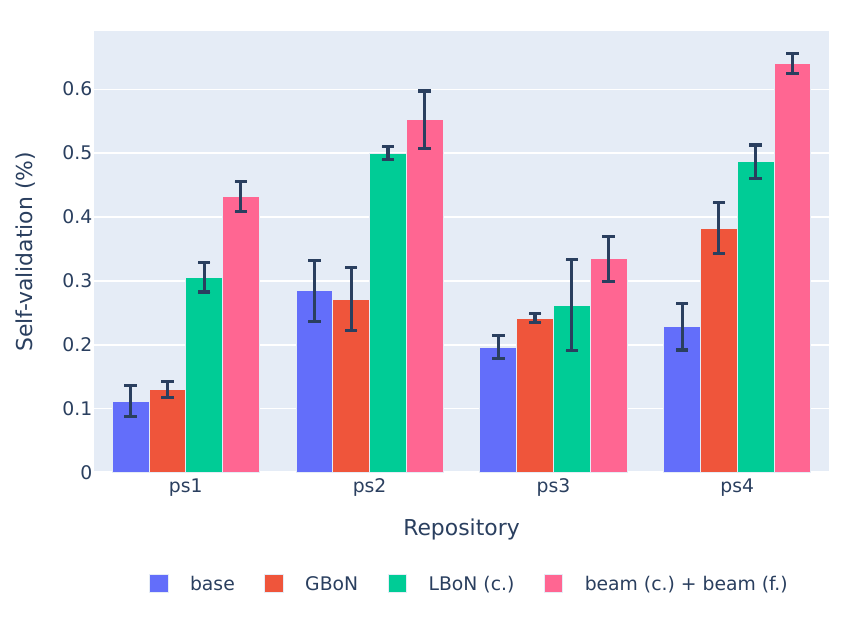}}
    \caption{}
    \label{fig:code-trans-all}
    \end{subfigure}
\caption{Results of using \oursimpl{} to apply
different inference-time scaling methods to the code repository translation agent.
All error bars show standard errors of the mean over 5 runs.
(a) A comprehensive hyperparameter search for \texttt{ps0};
(b) For \texttt{ps1} to \texttt{ps4},
we applying global best-of-$N$ (``GBoN''),
file-level local best-of-$N$ (``LBoN (c.)''),
and beam search at the file and method level (``beam (c.) + beam (f.)'')
while controlling for cost.
}
\label{fig:code-translation}
\end{center}
\vskip -0.2in
\end{figure}

\paragraph{Evaluation results}
\Cref{fig:code-trans-ps0} shows a log-linear plot of the scaling of various
inference-time strategies on \texttt{ps0}.
Consistent with prior work on inference-time scaling \citep{brown2024bon,li2022alphacode},
we find that performance scales linearly with the logarithm of the cost (all $\chi^2$ $p$-values $> 0.3$).
The best scaling is achieved with beam search applied at both the file level
and the individual method level (``beam (coarse), beam (fine)''),
outperforming the second best strategy ``beam (coarse)'' with a $p$-value of $0.2$
and all other strategies with statistical significance ($p < 0.03$).

Notably, the best-performing strategy  (``beam (coarse), beam (fine)'')
also happens to be the most difficult one to implement 
in plain Python.
It requires the programmer to break up the entire workflow
into all the individual LLM-calling steps where each step
explicitly stores and retrieves variables from a \lstinline|frame| dictionary.
This finding further demonstrates the merits of having a framework like \oursimpl{} where
experimenting with different search strategies can be done via simply changing a few parameters.
Combinations of agent and inference-time strategy
that have better scaling but that programmers would otherwise
choose not to implement due to their complexity of implementation,
are now made possible by \oursimpl{}.

We then evaluated the best-performing strategy ``beam (coarse), beam (fine)''
on \texttt{ps1} through \texttt{ps4} and compared it with
two simpler baselines
(``global best-of-$N$'', ``local best-of-$N$'')
while controlling for cost.
For beam search, we used a
file-level beam width of $2$
and a method-level beam width of $3$,
whereas we used $N = 16$ for both global and local best-of-$N$.
The average cost of a run was \$20--\$20.5 for \texttt{ps1},
\$27--\$30 for \texttt{ps2}, \$36--\$39 for \texttt{ps3}, and \$13.5--\$14
for \texttt{ps4}. The results are shown in \Cref{fig:code-trans-all}.
Overall, ``beam (coarse), beam (fine)'' continues to outperform
the other two simpler strategies.

To conclude, we have demonstrated the advantages of the separation of concerns
offered by \oursimpl{}. Implementing an inference-time strategy in \oursimpl{}
mainly involves adding branchpoints before LLM calls,
whereas without \oursimpl{}, significant source code modification that obscures the underlying
workflow is often necessary.
Furthermore, experimenting with different inference-time strategies in \oursimpl{}
is often as simple as changing a few search parameters.

\section{Related work}

\paragraph{Inference-time strategies for LLMs and agents}
\citep{welleck2024metageneration} provides a comprehensive review of algorithms
used during LLM inference to improve its reliability and performance.
Examples include best-of-$N$ sampling \citep{cobbe2021bon,brown2024bon,li2022alphacode}, refinement \citep{madaan2024selfrefine,shinn2024reflexion}, self-consistency \citep{wang2023selfconsistency}, and tree search 
\citep{yao2024tot,xie2024beam,yang2024leandojo},
which are also commonly used in
LLM-based agents \citep{zhou2023lats,koh2024agenttreesearch,antoniades2025swesearch,yamada2025aiscientistv2}.
\Cref{sec:agent_patterns} demonstrates how \oursimpl{} unifies and generalizes these inference-time scaling strategies for agents.

\paragraph{AI agent frameworks}
Several LLM-based agent frameworks have been developed
to abstract away boilerplate code and other low-level concerns,
and provide abstractions for common agentic patterns and components. 
AutoGen \citep{wu2024autogen} simplifies multi-agent conversation workflows with tool use,
LangChain \citep{chase2025langchain} simplifies linear workflows with RAG and tool use,
LangGraph \citep{campos2025langgraph} simplifies the creation of agent workflows as state machines,
and DSPy \citep{khattab2024dspy} automates prompt engineering.
Complementary to these efforts, our framework, \oursimpl{},
simplifies applying inference-time scaling strategies to agents.
Since \oursimpl{} involves adding statements such as \lstinline|branchpoint()|
to an existing agent written in Python,
it can be flexibly incorporated into
agents built with an existing Python agent framework.

\paragraph{Angelic nondeterminism}
Previous implementations of angelic nondeterminism
include John McCarthy's \lstinline|amb| operator in Common Lisp \citep{mccarthy1963amb}
and the list monad in Haskell \citep{wadler1995monads}.
The main conceptual difference is that 
\oursimpl{} implements a \textit{probabilistic} form of angelic nondeterminism,
which samples from a probability distribution such as an LLM
instead of choosing from a given set of choices.

\paragraph{Probabilistic programming}
Our work is also inspired by \textit{probabilistic programming},
a programming paradigm that separates the two main concerns of probabilistic inference:
specifying the probabilistic model and implementing the inference algorithm.
(See, e.g., \citep{goodman2013ppl} for a review.)
This allows the programmer to efficiently specify a probabilistic model in code
while independently experiment with different probabilistic inference algorithms.
Similarly, \oursimpl{} aims to separate the two main concerns of agent programming:
specifying the core agent workflow and implementing the inference-time search strategy.

\section{Limitations}

\oursimpl{} targets program-in-control style agents,
where implementations without \oursimpl{}
typically force the programmer to entangle the underlying agent
and the overlaying search strategy.
\oursimpl{} is not meant for
LLM-in-control style agents, where the two aspects are already decoupled.
Nevertheless, there has been increased interest in ``LLM+program-in-control''
hybrid style agents which involve an LLM writing a program-in-control style agent
\citep{li24coc,hu2025adas,yin2025godel}.
It would be interesting to explore using \oursimpl{} to make it easier for the LLM to
implement inference-time strategies in LLM-calling programs that it writes.

Although \oursimpl{} simplifies the source code modifications
needed to apply inference-time strategies to an existing agent,
modifications are still needed. There remains the engineering challenge
of choosing the correct places to add branchpoints, adding
sufficient and good-quality intermediate reward/verification signal,
and designing a good search algorithm.
\oursimpl{} could be improved to eliminate the need for source code modifications entirely,
where it solves the the majority of these remaining challenges by potentially using a flexible LLM-based search strategy.

\section{Conclusion}



This work introduced the \oursimpl{} programming framework,
which decouples the two fundamental aspects of agent programming:
defining the core agent workflow and designing the inference-time scaling strategy.
By enabling the integration of sophisticated search strategies into complex agent workflows,
\oursimpl{} opens up new possibilities for inference-time scaling of AI agents.
Looking ahead, we anticipate that the ability to seamlessly combine agent workflows
with powerful search techniques --- enabled by \oursimpl{} ---
will unlock new scaling laws
and drive the development of reliable LLM-augmented systems
for solving complex real-world tasks.




\bibliography{refs}
\bibliographystyle{unsrt}


\newpage
\section*{NeurIPS Paper Checklist}

\begin{enumerate}

\item {\bf Claims}
    \item[] Question: Do the main claims made in the abstract and introduction accurately reflect the paper's contributions and scope?
    \item[] Answer: \answerYes{}
    \item[] Justification: The abstract summarizes the problem and the proposed solution; the introduction defines the scope, identifies the problem, and summarizes the proposed solution and contributions.

\item {\bf Limitations}
    \item[] Question: Does the paper discuss the limitations of the work performed by the authors?
    \item[] Answer: \answerYes{}
    \item[] Justification: See Limitations section.

\item {\bf Theory assumptions and proofs}
    \item[] Question: For each theoretical result, does the paper provide the full set of assumptions and a complete (and correct) proof?
    \item[] Answer: \answerNA{}
    \item[] Justification: The paper proposes a programming framework and makes no theoretical claims.

    \item {\bf Experimental result reproducibility}
    \item[] Question: Does the paper fully disclose all the information needed to reproduce the main experimental results of the paper to the extent that it affects the main claims and/or conclusions of the paper (regardless of whether the code and data are provided or not)?
    \item[] Answer: \answerYes{}
    \item[] Justification: Compiler details are in \Cref{app:compiler_details}. Case studies describe experimental setup. Agent programs of case studies are in \Cref{app:plain-python}.

\item {\bf Open access to data and code}
    \item[] Question: Does the paper provide open access to the data and code, with sufficient instructions to faithfully reproduce the main experimental results, as described in supplemental material?
    \item[] Answer: \answerNo{}
    \item[] Justification: Company code

\item {\bf Experimental setting/details}
    \item[] Question: Does the paper specify all the training and test details (e.g., data splits, hyperparameters, how they were chosen, type of optimizer, etc.) necessary to understand the results?
    \item[] Answer: \answerYes{}
    \item[] Justification: See case study sections.

\item {\bf Experiment statistical significance}
    \item[] Question: Does the paper report error bars suitably and correctly defined or other appropriate information about the statistical significance of the experiments?
    \item[] Answer: \answerYes{}
    \item[] Justification: See case study sections, including tables and figures.

\item {\bf Experiments compute resources}
    \item[] Question: For each experiment, does the paper provide sufficient information on the computer resources (type of compute workers, memory, time of execution) needed to reproduce the experiments?
    \item[] Answer: \answerYes{}
    \item[] Justification: Case study sections describe LLM costs and CPU used.
    
\item {\bf Code of ethics}
    \item[] Question: Does the research conducted in the paper conform, in every respect, with the NeurIPS Code of Ethics \url{https://neurips.cc/public/EthicsGuidelines}?
    \item[] Answer: \answerYes{}
    \item[] Justification: No aspect of the research violated the Code of Ethics.

\item {\bf Broader impacts}
    \item[] Question: Does the paper discuss both potential positive societal impacts and negative societal impacts of the work performed?
    \item[] Answer: \answerNA{}
    \item[] Justification: Foundational research with standard expectations in regards to societal impact
    
\item {\bf Safeguards}
    \item[] Question: Does the paper describe safeguards that have been put in place for responsible release of data or models that have a high risk for misuse (e.g., pretrained language models, image generators, or scraped datasets)?
    \item[] Answer: \answerNA{}
    \item[] Justification: No release of data/models with a high risk for misuse

\item {\bf Licenses for existing assets}
    \item[] Question: Are the creators or original owners of assets (e.g., code, data, models), used in the paper, properly credited and are the license and terms of use explicitly mentioned and properly respected?
    \item[] Answer: \answerYes{}
    \item[] Justification: Existing agents and benchmarks cited and license terms followed

\item {\bf New assets}
    \item[] Question: Are new assets introduced in the paper well documented and is the documentation provided alongside the assets?
    \item[] Answer: \answerNA{}
    \item[] Justification: No new assets released.

\item {\bf Crowdsourcing and research with human subjects}
    \item[] Question: For crowdsourcing experiments and research with human subjects, does the paper include the full text of instructions given to participants and screenshots, if applicable, as well as details about compensation (if any)? 
    \item[] Answer: \answerNA{}
    \item[] Justification: No human subjects involved.

\item {\bf Institutional review board (IRB) approvals or equivalent for research with human subjects}
    \item[] Question: Does the paper describe potential risks incurred by study participants, whether such risks were disclosed to the subjects, and whether Institutional Review Board (IRB) approvals (or an equivalent approval/review based on the requirements of your country or institution) were obtained?
    \item[] Answer: \answerNA{}
    \item[] Justification: No human subjects involved.

\item {\bf Declaration of LLM usage}
    \item[] Question: Does the paper describe the usage of LLMs if it is an important, original, or non-standard component of the core methods in this research? Note that if the LLM is used only for writing, editing, or formatting purposes and does not impact the core methodology, scientific rigorousness, or originality of the research, declaration is not required.
    \item[] Answer: \answerNA{}
    \item[] Justification: LLMs were not used to help produce this research other than as a coding and writing tool.

\end{enumerate}


\newpage

\appendix

\section{Additional case studies} \label{app:case_studies}

This appendix presents Case Studies 2 and 3. In these case studies,
we study agents much simpler than
the code translation agent in our main case study (Case Study 1)
so that we can more explicitly compare code written in \oursimpl{} vs.\ plain Python.
Our objective is to illustrate and understand how the modularity that \oursimpl{} provides
lets programmers more easily implement and experiment with different inference-time
scaling strategies for their agent.

Experiments for all case studies were conducted on a Macbook Pro
with an M3 chip and 18\,GB of RAM.
All LLM calls were made through the OpenAI API.

\subsection{Case Study 2: Hypothesis Search Agent} \label{sec:hypo_search}

In this case study, we use a simple two-step agent for ARC-AGI \citep{chollet2019arc}
to illustrate how \oursimpl{} enables the programmer to quickly implement inference-time search.

\paragraph{Base agent}
A task in ARC-AGI shows the agent around 3 validation examples
of input-output grid pairs,
and the objective is to find the rule that transforms input grids into output grids
and apply the rule to a test input grid.
A simple agent for solving ARC-AGI tasks is as follows (\Cref{lst:arc_baseline}):
1.\ ask the LLM for a natural language hypothesis of the transformation rule;
2.\ ask the LLM to implement the hypothesis in code.
\begin{lstlisting}[basicstyle=\ttfamily\small, caption=Simple 2-step agent for ARC, label=lst:arc_baseline]
def two_step_agent(task_info):
    # Step 1: Get natural language hypothesis
    ...
    hypothesis = hypothesis_agent([task_info], hypothesis_instruction)

    # Step 2: Implement the hypothesis in code
    ...
    code = solver_agent([task_info, hypothesis], solver_instruction)
    return get_test_output(code)
\end{lstlisting}

\paragraph{The \oursimpl{} agent}
To convert this agent into a \oursimpl{} program,
we identify the points of unreliability: the two LLM calls.
Before each LLM call, we can add a branchpoint to allow the external search algorithm
to search over different samples from the LLM.
Finally, we add a final verification step that evaluates the generated code on the validation grid pairs,
so that the search algorithm knows which execution paths did better.
Here is the resulting \oursimpl{} agent (\Cref{lst:hypo_encompass}):
\begin{lstlisting}[basicstyle=\ttfamily\small, caption=Two-step agent with BFS in \oursimpl{} reproduces Hypothesis Search, label=lst:hypo_encompass]
@encompass.compile
def two_step_agent(task_info):
    # 1st branchpoint results in multiple samples of the natural language hypothesis
    branchpoint()
    # Step 1: Get natural language hypothesis
    ...
    hypothesis = hypothesis_agent([task_info], hypothesis_instruction)

    # 2nd branchpoint results in multiple code samples for each hypothesis
    branchpoint()
    # Step 2: Implement the hypothesis in code
    ...
    code = solver_agent([task_info, hypothesis], solver_instruction)

    # Evaluate
    percent_correct, feedback = run_validation(code)
    record_score(n_correct)
    if percent_correct == 1.0:
        early_stop_search()

    return get_test_output(code)

two_step_agent(task_info).search("parallel_bfs", default_branching=8)
\end{lstlisting}
Here, we've chosen 8 samples of subsequent execution from each branchpoint
and apply parallelized breadth-first search (parallel BFS) over all program execution paths,
In particular, BFS samples 8 natural language hypotheses following the first branchpoint,
and for each hypothesis samples 8 code implementations from the second branchpoint.
It then chooses the result from the 64 implementations with the highest evaluation score (recorded by \lstinline|record_score|).
This replicates a version of Hypothesis Search \citep{wang2023hypothesis}
without the hypothesis summarization step and execution feedback loop.

We also consider an agent with only the first of the two branchpoints.
This gives rise to global best-of-$N$ sampling, i.e.,
running the base agent $N$ times in parallel and keeping the best run.

\paragraph{Comparison with equivalent plain Python}
In implementing the \oursimpl{} agent,
because the changes made to the original agent are minimal,
the underlying logic of the agent is clearly portrayed by the code,
with the external search logic (sampling) indicated by a few branchpoint statements.

We now compare this with the equivalent agent in plain Python.
For the one-branchpoint hypothesis search agent (best-of-$N$), it is still
relatively straightforward to implement it in plain Python by running $N$ copies of the agent
in $N$ parallel threads until we find a solution that passes validation.

However, to further add the second branchpoint --- which is just an additional line of code
in EnCompass --- the equivalent implementation in plain Python of parallel BFS
requires significant structural changes. In defining the tasks to be executed in
a multithreaded fashion, the underlying agent workflow has been broken up
and the program flow obscured,
even though the agent only contains two steps (\Cref{lst:hypothesis_parallel_main}).
\begin{lstlisting}[basicstyle=\ttfamily\scriptsize, caption={Parallelized BFS in plain Python, obscuring the underlying two-step agent workflow}, label=lst:hypothesis_parallel_main]
from concurrent.futures import ThreadPoolExecutor, as_completed


def two_step_agent(task_info, branching):
    results = []
    full_solved = False

    with ThreadPoolExecutor() as executor:

        def run_one_forward_pass():
            if full_solved:
                return
            # Step 1: Get natural language hypothesis
            ...
            hypothesis = hypothesis_agent([task_info], hypothesis_instruction)

            def implement_in_code():
                nonlocal full_solved

                if full_solved:
                    return

                # Step 2: Implement the hypothesis in code
                ...
                code = solver_agent([task_info, hypothesis], solver_instruction)

                # Evaluate
                percent_correct = run_validation(code)
                if percent_correct == 1:
                    full_solved = True
                results.append((get_test_output(code), percent_correct))

            futures = [executor.submit(implement_in_code) for _ in range(branching)]
            for future in as_completed(futures):
                future.result()

        futures = [executor.submit(run_one_forward_pass) for _ in range(branching)]
        for future in as_completed(futures):
            future.result()

    return max(results, key=lambda x: x[1])[0]


two_step_agent(task_info, branching=8)
\end{lstlisting}


\newpage

\paragraph{Evaluation}
The purpose of this evaluation section is to complete our demonstration
of using \oursimpl{} to implement and compare different inference-time scaling strategies. 

We use a subset of the ARC-AGI benchmark corresponding to the 60 tasks
sampled from the ``Public Training Set (Easy)'' that ADAS \citep{hu2025adas} used.
We report the mean evaluation score as well as its standard error over 5 seeds.

We evaluate the following agents on this ARC-AGI subset:
\begin{itemize}
    \item The two-step agent (base agent)
    \item Global best-of-$N$ applied to the two-step agent (one branchpoint),
    where $N = 8, 36$
    \item The Hypothesis Search agent \citep{wang2023hypothesis},
    i.e., parallelized BFS applied to the two-step agent (two branchpoints),
    with branching factor $8$.
\end{itemize}
The LLM temperature was set to $0.8$ for all experiments.

The evaluation results are shown in \Cref{tab:arc}.
The results show how scaling inference-time compute
by adding \lstinline|branchpoint()| statements and adjusting search parameters
quickly increases the evaluation accuracy to results better than 
the best agent discovered by costly meta-agent search (ADAS) \citep{hu2025adas}.
Comparing the two scaling strategies, we find that best-of-$N$ and BFS
are comparable. 



\begin{table}[tb]
\caption{Percentage accuracy of a simple two-step agent on a subset of ARC with progressively more
\oursimpl{} branchpoints:
no branchpoints, 1 branchpoint at the top, and 2 branchpoints before the 2 LLM calls.
Accuracy improves quickly as more branchpoints are added.
We also compare with the best agent discovered through meta-agent search (ADAS \citep{hu2025adas}).}
\label{tab:arc}
\vskip 0.15in
\begin{center}
\begin{small}
\begin{tabular}{rcccc}
\toprule
Base model & \multicolumn{2}{c}{GPT-3.5} & \multicolumn{2}{c}{GPT-4o} \\
& Acc.\ (\%) & Total cost & Acc.\ (\%) & Total cost \\
\midrule
Two-step agent   & 4.3 $\pm$ 0.9 & \$0.41 & 24.0 $\pm$ 1.5 & \$2.85 \\
\rev{+ global best-of-$N$, $N = 8$ (ours)}  & 11.7 $\pm$ 0.8 & \$3.29 & 36.3 $\pm$ 1.1 & \$22.76 \\
\rev{+ global best-of-$N$, $N = 36$ (ours)}  & \rev{\textbf{16.0} $\pm$ 1.0} & \$14.81 & \rev{\textbf{38.7} $\pm$ 1.1} & \$95.98 \\
\rev{+ BFS, branching $= 8$ (ours)} & \textbf{15.0} $\pm$ 0.9 & \$15.81 & \textbf{38.3} $\pm$ 1.2 & \$88.69 \\
\midrule
ADAS best agent (reported)$^\dagger$ & 13.7 $\pm$ 2.0 & --- & 30.0 $\pm$ 2.6 & --- \\
ADAS best agent (reproduced) & 10.7 $\pm$ 0.8 & \$2.11 & 32.7 $\pm$ 1.1 & \$27.85 \\
\bottomrule
\end{tabular}
\end{small}
\end{center}
\scriptsize
$^\dagger$The reported results use a different checkpoint of GPT-4o and the errors are estimated differently,
using a bootstrapping confidence interval.
\vskip -0.1in
\end{table}

\subsection{Case Study 3: Reflexion Agent} \label{sec:reflexion}

In this case study, we show how applying \oursimpl{} to an existing
agentic pattern provides a new dimension for the cost-efficient scaling of inference-time compute.

\paragraph{Base agent}
As our baseline, we use Reflexion \citep{shinn2024reflexion}
as a coding agent (\Cref{lst:reflexion_base}),
which uses an LLM to iteratively reflect on past attempts and their feedback to improve the response.
Feedback includes both LLM-generated self-reflection and
results from running LLM-generated unit tests.

\paragraph{The \oursimpl{} agent}
In \oursimpl{}, we modify Reflexion by adding two branchpoints (\Cref{lst:reflexion_ours}): one before the initial code generation,
and one at the top of the body of the for loop (i.e., before each iteration of self-reflection plus generation).
Pass rate on the LLM-generated unit tests feedback is
used as the verification score in \lstinline|record_score()|
in the \oursimpl{} agent.
We apply two search strategies:
one is global best-of-$N$, and the other one is
``reexpand best-first search'', our variant of best-first search (BeFS)
where the strategy is to simply always
choose the node with the highest verification score to step.

\paragraph{Comparison with equivalent plain Python}
Implementing best-of-$N$ sampling in plain Python is straightforward ---
simply wrap the agent in a for loop.
However, to implement reexpand best-first search in the Reflexion agent,
a plain Python implementation
requires structural changes to the code
when \oursimpl{} only requires adding two branchpoints.
In particular, the initial sampling step and the self-reflection step are put into separate functions,
corresponding to the 2 actions that the agent is allowed to take (\Cref{lst:reflexion_plain} in \Cref{app:python_case2}).
The agent maintains a search tree and iteratively chooses the best node to expand:
if the chosen node is the root node, then a new code sample is drawn from the LLM,
whereas if the node is not the root, then a self-reflection step is applied to it.

Furthermore, separating the two actions into separate functions
loses the natural logical ordering between them (the initial sampling step
should occur before the self-reflection step).
For more complex agent workflows like the code repository translation agent in
\Cref{sec:codetrans},
the original underlying agent workflow becomes heavily obscured.

\paragraph{Evaluation}
The purpose of this evaluation section is to complete our demonstration
of using \oursimpl{} to implement and compare different inference-time scaling strategies.

LeetCode is a website with programming exercises to help prepare for software engineer interviews,
and the LeetCodeHard benchmark is a collection of 40 hard LeetCode problems \citep{shinn2024reflexion}.
A problem typically has a few dozen test cases (occasionally a few hundred or over a thousand test cases).
While the LLM agent does not see these test cases, it can use LLM-generated test cases.
We calculate the evaluation score as the average pass rate over all 40 problems,
where the pass rate for any given problem is the fraction of test cases passed.

For both the base agent and the \oursimpl{} BeFS agent,
we consider 3 different cost settings (low, medium, high)
where the number of code generations $n = 5, 8, 13$.
In the base agent, we vary the number of feedback loops to be $4, 7, 12$,
whereas for the BeFS agent, the number of feedback loops is fixed at $4$
but the total number of code generations is controlled by the
external search algorithm algorithm.
In the best-of-$N$ agent, we have 2 cost settings (low, high)
by adjusting $N = 1, 2$.
The temperature of the LLM is set to $0.0$ in the base agent
and $0.5$ ($n = 5, 8$) or $1.0$ ($n = 13$) in the \oursimpl{} agent.

As shown in \Cref{tab:reflexion}, controlling inference-time scaling through the
external search algorithm in \oursimpl{} scales in a more cost-efficient manner
than scaling the number of feedback loops in Reflexion:
the same performance is achieved at a lower cost.
Comparing the two scaling strategies, we find that BeFS and best-of-$N$
are comparable. 

\begin{table*}[ht]
\vskip -0.05in
\caption{Increasing the number of search steps in the \oursimpl{} Reflexion agent
scales better than scaling the number of refinement loops in the vanilla Reflexion agent:
the same performance is achieved at a lower cost. All errors are standard errors of the mean over 5 runs.}
\label{tab:reflexion}
\begin{center}
\vskip -0.1in
\begin{small}
\begin{tabular}{rcccccc}
    \toprule
    Cost setting & \multicolumn{2}{c}{Low} & \multicolumn{2}{c}{Medium} & \multicolumn{2}{c}{High} \\
    \cmidrule(lr){2-3} \cmidrule(lr){4-5} \cmidrule(lr){6-7}
    & Acc.\ (\%) & Cost/task (\$) & Acc.\ (\%) & Cost/task (\$) & Acc.\ (\%) & Cost/task (\$) \\
    \midrule
    Reflexion & 35.5 $\pm$ 1.0 & 0.279 $\pm$ 0.005 & 35.9 $\pm$ 1.3 & 0.449 $\pm$ 0.005 & 38.2 $\pm$ 1.2 & 0.736 $\pm$ 0.010 \\
    +best-of-$N$ & 35.5 $\pm$ 1.0 & 0.279 $\pm$ 0.005 & --- & --- & 37.6 $\pm$ 1.7 & \textbf{0.508} $\pm$ 0.013 \\
    +BeFS & 36.1 $\pm$ 2.1 & \textbf{0.168} $\pm$ 0.004 & 36.1 $\pm$ 1.1 & \textbf{0.289} $\pm$ 0.007 & 38.1 $\pm$ 1.3 & \textbf{0.512} $\pm$ 0.006 \\
    \bottomrule
\end{tabular}
\end{small}
\end{center}
\vskip -0.1in
\end{table*}

\newpage

\section{Documentation of \oursimpl{}} \label{app:documentation}

\oursimpl{} is an instantiation of the \ours{} programming framework in Python.
It is implemented as the \lstinline|@encompass.compile| function decorator,
which makes several new keywords primitives available in the body of the decorated function.
\Cref{app:compiler_details} describes how the decorator compiles the function body
into an object that provides an interface for search.

This appendix is organized as follows:
\begin{itemize}
    \item 
    \Cref{app:primitives} lists all \oursimpl{} keyword primitives that are made available
    inside a function with the \lstinline|@encompass.compile| decorator.

    \item
    \Cref{app:compiled} describes the interface of the compiled search space object
    created by the \lstinline|@encompass.compile| decorator.

    \item
    \Cref{app:checkpoint} describes the interface of the \lstinline|Checkpoint| object
    that represents the program state at a branchpoint or return statement.

    \item
    \Cref{app:search} describes the search algorithms that \oursimpl{} provides out-of-the-box,
    as well as the abstract \lstinline|Search| class that the user can subclass
    to define their own custom search algorithms.
\end{itemize}

\subsection{\oursimpl{} primitives} \label{app:primitives}

The following is the complete list of the 12 \oursimpl{} keyword primitives in alphabetical order.
They are available in any function or async function with the \lstinline|@encompass.compile| decorator.

\lstinline[emph={[2]branchpoint_params}, emphstyle={[2]\itshape}]|branchpoint(**branchpoint_params)|

\begin{documentation}
    This statement marks a \textit{branchpoint}. When combined with proper verification signal
    from \lstinline|record_score| statements (see below), this creates the illusion
    that the stochastic operations that follow are now biased to more desirable outputs,
    and unreliable operations (e.g., LLM calls) have become more reliable.

    This illusion (\textit{angelic nondeterminism}) is accomplished through
    search over the different nondeterministic branches of the program's execution.
    More specifically, when the program's execution reaches a branchpoint,
    the program will branch into multiple copies of itself and
    an external search algorithm implemented using the \lstinline|Checkpoint|
    interface searches over the multiple branches of the program.

    \lstinline[emph={[2]branchpoint_params}, emphstyle={[2]\itshape}]|branchpoint_params|
    can include the following keyword arguments (all are optional):
    \begin{itemize}
        \item \lstinline|name: Any|: A name to label the branchpoint
        \item \lstinline!max_protection: int | None!: The maximum number of times stepping to the next branchpoint is allowed to raise an exception that gets protected (see documentation for \lstinline|protect()|).
        \item \lstinline|message_to_agent: Any|: A message to send to the agent (see below for messaging).
    \end{itemize}
    Other available keyword arguments depend on the specific search algorithm being used.
    For example, algorithms derived from graph search algorithms by fixing the branching factor allow the programmer to provide a branchpoint-specific branching factor \lstinline|branching| and maximum amount of parallelization \lstinline|max_workers| when sampling the next state.

    \textit{Example usage:} The simplest use case is to add one \lstinline|branchpoint()|
    statement at the top of the function body (Listing \ref{lst:sampling}), which amounts to best-of-$N$ sampling (\Cref{sec:bestofn}):

    \begin{lstlisting}[basicstyle=\ttfamily\small, caption=\lstinline|branchpoint| example: Best-of-$N$ sampling, label=lst:sampling]
@encompass.compile
def branchpoint_example(...):
    branchpoint()
    ...  # Do something
    record_score(...)

# Sample 10 times and output the result with the highest score
branchpoint_example(...).search("sampling", num_rollouts=10)
    \end{lstlisting}

    \lstinline|branchpoint()| also supports messaging with the controller
    (user of the \lstinline|Checkpoint| interface) with a syntax similar to that of
    Python \lstinline|yield|. This lets the programmer implement highly customized
    search algorithms optimized for their particular agent workflow ---
    decisions on node selection and backtracking
    can now depend on the details of the execution state of the agent
    that are sent to the search process via this messaging interface.

    \textit{Example usage:} \Cref{lst:messaging} illustrates how messaging can be used
    to let the controller decide whether to backtrack based on the execution state
    of the underlying agent.
\begin{lstlisting}[basicstyle=\ttfamily\small, caption=Example of \lstinline|branchpoint| with agent-controller messaging, label=lst:messaging]
@encompass.compile
def branchpoint_messaging(task):
    branchpoint()
    solution = ...
    feedback = ...
    # Python equivalent: response = yield (...)
    response = branchpoint(message_to_controller=(task, solution, feedback))
    print(response)

# Python equivalent: generator = branchpoint_messaging(); next(generator)
checkpoint0 = branchpoint_messaging().start()
# Python equivalent: task, solution, feedback = next(generator)
checkpoint1 = checkpoint0.step()
task, solution, feedback = checkpoint1.message_from_agent
# Decide whether to backtrack
should_backtrack = decide_backtrack(task, solution, feedback)
if should_backtrack:
    # Backtrack and retry last step - no Python equivalent
    checkpoint1 = checkpoint0.step()
# Python equivalent: generator.send(f"backtracked: {should_backtrack}")
checkpoint2 = checkpoint1.step(
    message_to_agent=f"backtracked: {should_backtrack}"
)
\end{lstlisting}

\end{documentation}

\lstinline[emph={[2]choices, branchpoint_params}, emphstyle={[2]\itshape}]|branchpoint_choose(choices: Iterable, **branchpoint_params)|:

\begin{documentation}
    This is a variant of \lstinline|branchpoint| where the resulting branches have the 
    \lstinline|branchpoint_choose(choices)| expression evaluate to the elements
    in the iterable \lstinline|choices|.
    In other words, this implements regular angelic nondeterminism.

    \textit{Example usage:} The following function (Listing \ref{lst:graphsearch}) guesses a path
    from a start node to a goal in a graph. Conducting search over the nondeterministic execution branches
    becomes equivalent to actual search over the graph.

    \begin{lstlisting}[basicstyle=\ttfamily\small, caption=Graph search example with \lstinline|branchpoint_choose|, label=lst:graphsearch]
@encompass.compile
def graph_search(graph, start_node, goal):
    """
    Guess a path from `start_node` to `goal` in a `graph` represented as an adjacency list.
    """
    cur_node = start_node
    path = [cur_node]
    cost_so_far = 0
    while cur_node != goal:
        next_node = branchpoint_choose(graph[cur_node], identity=cur_node)
        path = path + [cur_node]
        cost_so_far += get_edge_cost(cur_node, next_node)
        total_estimated_cost = cost_so_far + estimate_cost_to_go(next_node, goal)
        record_score(-total_estimated_cost)
        cur_node = next_node
    return path

# Conduct best-first search -> shortest path with A* search
graph_search(my_graph, my_start_node, my_goal).search("best_first", top_k_popped=1, default_branching=None)
    \end{lstlisting}
\end{documentation}

\lstinline|early_stop_search()|

\begin{documentation}
    This early-stops the external search process because, e.g., a correct answer has been found.

    \textit{Example usage:} (also see Case Studies 2 and 3)
    \begin{lstlisting}[basicstyle=\ttfamily\small, caption=\lstinline|early_stop_search| example, label=lst:early_stop_search]
@encompass.compile
def early_stop_search_example(...):
    ...  # Do something before
    branchpoint()
    # Ask LLM to generate answer
    answer = llm.generate(...)
    # Check answer
    success = check_answer(answer)
    if success:
        early_stop_search()
    return answer
    \end{lstlisting}
\end{documentation}

\lstinline[emph={[2]err}, emphstyle={[2]\itshape}]|kill_branch(err=None)|

\begin{documentation}
    This kills the current branch of program execution.
    For example, if the LLM generated something irreparably bad,
    instead of recording a large negative score (i.e., \lstinline|record_score(-1000)|),
    one can simply kill the current branch.

    \textit{Example usage:}
    \begin{lstlisting}[basicstyle=\ttfamily\small, caption=Example usage of \lstinline|kill_branch|, label=lst:kill_branch]
@encompass.compile
def kill_branch_example(...):
    ...  # Do something before
    branchpoint()
    # Ask LLM to do something
    response = llm.generate(...)
    sanity_check_passed = sanity_check_llm_response(response)
    if not sanity_check_passed:
        kill_branch()
    ...  # Do something after
    \end{lstlisting}
\end{documentation}

\lstinline[emph={[2]var, expr}, emphstyle={[2]\itshape}]|var: NeedsCopy      var: NeedsCopy = expr|

\begin{documentation}
    This tells the \oursimpl{} compiler that the variable named
    \lstinline[emph={[2]var}, emphstyle={[2]\itshape}]|var|
    needs to be copied upon branching.
    In other words, this type annotation declares a variable
    that is independent across all future execution paths of the program,
    assuming no
    ``\lstinline[emph={[2]var}, emphstyle={[2]\itshape}]|var: NoCopy|''
    declaration ever occurs in the future.

    By default, all local variables need copying,
    so \lstinline|NeedsCopy| is typically only used to
    undo an earlier \lstinline|NoCopy| declaration.

    Global variables are never copied. In fact,
    using ``\lstinline[emph={[2]var}, emphstyle={[2]\itshape}]|var: NeedsCopy|''
    in a Python function will actually declare a \textit{local} variable
    named \lstinline[emph={[2]var}, emphstyle={[2]\itshape}]|var|
    that needs copying.

    Note that variable assignment without a \lstinline|NeedsCopy|
    or \lstinline|NoCopy| declaration will not change
    whether it is \lstinline|NeedsCopy| or \lstinline|NoCopy|.

    \textit{Example usage:} In this example,
    the programmer wishes to reuse the name of a \lstinline|NoCopy|
    variable for something that needs copying (\Cref{lst:needs_copy}):
    
\begin{lstlisting}[basicstyle=\ttfamily\small, caption=\lstinline|NeedsCopy| example, label=lst:needs_copy]
@encompass.compile
def needs_copy_example(task):
    # Step 1: Iterative refinement using NoCopy
    feedbacks: NoCopy = []
    branchpoint()
    ...
    score, feedback = get_score_and_feedback(...)
    feedbacks.append(feedback)
    record_score(score)

    # Step 2: Summarize every feedback in `feedbacks`
    feedbacks: NeedsCopy  # Different summary attempts mutate differently --- so we want copies of `feedbacks` on different search branches
    branchpoint()  # Sample multiple summary attempts
    for i, feedback in enumerate(feedbacks):
        feedbacks[i] = summarize_feedback(feedback)
    ...

result = agent_forward(task).search("dfs", default_braching=5)
    \end{lstlisting}
\end{documentation}

\lstinline[emph={[2]var, expr}, emphstyle={[2]\itshape}]|var: NoCopy      var: NoCopy = expr|

\begin{documentation}
    This tells the \oursimpl{} compiler that the variable named
    \lstinline[emph={[2]var}, emphstyle={[2]\itshape}]|var|
    need not be copied upon branching.
    In other words, this type annotation declares a variable
    that is shared across all future execution paths of the program,
    assuming no
    ``\lstinline[emph={[2]var}, emphstyle={[2]\itshape}]|var: NeedsCopy|''
    declaration ever occurs in the future.

    By default, all local variables need copying,
    so \lstinline|NoCopy| is needed to declare a variable
    to be shared across future execution paths.

    Global variables are never copied, so there is no need
    to use ``\lstinline[emph={[2]var}, emphstyle={[2]\itshape}]|var: NoCopy|''
    to specify a global variable that doesn't need copying.
    In fact, this declaration would actually
    declare a \textit{local} variable
    named \lstinline[emph={[2]var}, emphstyle={[2]\itshape}]|var|
    that doesn't need copying.

    Note that variable assignment without a \lstinline|NeedsCopy|
    or \lstinline|NoCopy| declaration will not change
    whether it is \lstinline|NeedsCopy| or \lstinline|NoCopy|.

    \textit{Example usage:} The simplest use case
    is to modify the best-of-$N$ (one branchpoint at the top)
    by initializing a shared memory of feedback from past attempts.
    This gives rise to iterative refinement (\Cref{sec:refinement}).
\begin{lstlisting}[basicstyle=\ttfamily\small, caption=Iterative refinement, label=lst:refine]
@encompass.compile
def no_copy_example(task):
    feedbacks: NoCopy = []
    branchpoint()
    result = perform_task(task, feedbacks)
    score, feedback = get_score_and_feedback(result)
    feedbacks.append(feedback)
    record_score(score)

# Sample 10 times and output the result with the highest score
result = agent_forward(task).search("sampling", num_rollouts=10)
\end{lstlisting}
\end{documentation}

\lstinline[emph={[2]return_value}, emphstyle={[2]\itshape}]|optional_return(return_value)|

\begin{documentation}
    This signals to the external search process that, although the program execution hasn't finished,
    an output \lstinline[emph={[2]return_value}, emphstyle={[2]\itshape}]|return_value|
    has already been produced and should be treated as a possible return value of the program.

    \textit{Example usage:} (also see Listing \ref{lst:reflexion_ours} in Case Study 2)

    \begin{lstlisting}[basicstyle=\ttfamily\small, caption=\lstinline|optional_return| answer, label=lst:optional_return]
@encompass.compile
def optional_return_example(...):
    answer = llm.generate_answer(...)
    optional_return(answer)
    refined_answer = llm.refine_answer(answer, ...)
    return refined_answer
    \end{lstlisting}
\end{documentation}

\lstinline[emph={[2]expr, exception, max_retries}, emphstyle={[2]\itshape}]|protect(expr, exception, max_retries=None)|

\begin{documentation}
    If evaluating an expression \lstinline[emph={[2]expr}, emphstyle={[2]\itshape}]|expr| may
    raise exception \lstinline[emph={[2]exception}, emphstyle={[2]\itshape}]|exception|,
    then wrapping it in \lstinline|protect(...)| creates the illusion that it no longer raises the exception.
    The illusion is created by resampling from the most recent branchpoint
    until evaluating the expression no longer raises the exception.
    \lstinline[emph={[2]max_retries}, emphstyle={[2]\itshape}]|max_retries|,
    if not \lstinline|None|, sets an upper limit on the number of retries.

    \textit{Example usage:} One example use case is parsing output from an LLM.
    The following example extracts the Python code block from an LLM and parses it.
    Both steps could error out because of the unreliability of the LLM,
    so we can wrap them in \lstinline|protect|.
    \begin{lstlisting}[basicstyle=\ttfamily\small, caption=\lstinline|protect| example: Safely parsing output from an LLM, label=lst:parse_llm_output]
@encompass.compile
def parse_llm_output_example(...):
    ...  # Do something before
    branchpoint()
    # Ask LLM to generate Python code
    response = llm.generate(...)
    # Extract Python code
    python_code = protect(response.split("```python\n", 1)[1]
                                  .split("```", 1)[0], IndexError)
    # Parse Python code
    python_ast = protect(ast.parse(python_code), SyntaxError)
    ...  # Do something after
    \end{lstlisting}
\end{documentation}

\lstinline[emph={[2]costs}, emphstyle={[2]\itshape}]|record_costs(**costs)|

\begin{documentation}
    This lets the user track various kinds of cost, e.g., LLM usage.
    The costs are aggregated and accessed through the dictionary
    \lstinline[emph={[2]func}, emphstyle={[2]\itshape}]|func.aggreagte_costs|
    where \lstinline[emph={[2]func}, emphstyle={[2]\itshape}]|func| is the compiled function.

    \textit{Example usage:}

    \begin{lstlisting}[basicstyle=\ttfamily\small, caption=\lstinline|record_costs| example, label=lst:record_costs]
@encompass.compile
def record_costs_example(...):
    response, cost = llm.generate(...)
    record_costs(llm_cost=cost, llm_num_calls=1)
    return response
    \end{lstlisting}
\end{documentation}

\lstinline[emph={[2]score}, emphstyle={[2]\itshape}]|record_score(score)|

\begin{documentation}
    This is the main means for providing reward/verification signal to the external search algorithm
    by recording a score.
    The exact semantics of this score will depend
    on the search algorithm used
    (e.g., heuristic for best-first search,
    value function for MCTS).

    \textit{Example usage:} The simplest example is best-of-$N$ sampling,
    which samples the agent workflow multiple times and selects the result
    with the highest score recorded by \lstinline|record_score|.

    \begin{lstlisting}[basicstyle=\ttfamily\small, caption=\lstinline|record_score| example: Best-of-$N$ sampling, label=lst:sampling2]
@encompass.compile
def branchpoint_example(...):
    branchpoint()
    ...  # Do something
    record_score(...)

# Sample 10 times and output the result with the highest score
branchpoint_example(...).search("dfs", default_branching=10)
    \end{lstlisting}
\end{documentation}

\lstinline[emph={[2]group_evaluator, eval_target, eval_label}, emphstyle={[2]\itshape}]|record_score(group_evaluator, eval_target, label=eval_label)|

\begin{documentation}
    This overloading of \lstinline|record_score| enables evaluation that compares across
    multiple program execution branches. The simplest use case for this is self-consistency majority voting,
    where evaluating a result must be done relative to all results (\Cref{sec:self-consistency}).
\end{documentation}

\lstinline[emph={[2]func, args, kwargs}, emphstyle={[2]\itshape}]|searchover(func(...))|

\begin{documentation}
    This is the syntax for calling an \oursimpl{} function \lstinline[emph={[2]func}, emphstyle={[2]\itshape
}]|func|
    inside another \oursimpl{} function.
    This is similar to the \lstinline[emph={[2]func}, emphstyle={[2]\itshape}]|await func(...)|
    syntax for calling an async function inside another async function,
    where instead of \lstinline|await| we use \lstinline|searchover|.

    \textit{Example usage:} (also see Listing \ref{lst:code_trans_ours} in Case Study 1)

    \begin{lstlisting}[basicstyle=\ttfamily\small, caption=\lstinline|searchover| example, label=lst:searchover]
@encompass.compile
def helper_function(...):
    ...

@encompass.compile
def searchover_example(...):
    ...  # Do something before
    helper_result = searchover(helper_function(...))
    ...  # Do something after
    \end{lstlisting}
\end{documentation}

\lstinline[emph={[2]async_func}, emphstyle={[2]\itshape}]|searchover_await(async_func(...))|

\begin{documentation}
    This is the asynchronous counterpart to \lstinline|searchover()|.
    In other words, it is used to call an asynchronous \oursimpl{} function
    \lstinline[emph={[2]async_func}, emphstyle={[2]\itshape}]|async_func|
    from within another asynchronous \oursimpl{} function.

    \textit{Example usage:}

    \begin{lstlisting}[basicstyle=\ttfamily\small, caption=\lstinline|searchover_await| example, label=lst:searchover_await]
@encompass.compile
async def async_helper_function(...):
    ...

@encompass.compile
async def searchover_await_example(...):
    ...  # Do something before
    helper_result = searchover_await(async_helper_function(...))
    ...  # Do something after
    \end{lstlisting}
\end{documentation}

\subsection{Compiled search space interface} \label{app:compiled}

The interface of the compiled search space allows the user to either \textit{step}
through the program or \textit{search} over its nondeterministic execution paths.

In what follows, \lstinline[emph={[2]func}, emphstyle={[2]\itshape}]|func|
represents a function compiled with the \lstinline|@encompass.compile| decorator,
and \lstinline[emph={[2]func}, emphstyle={[2]\itshape}]|func(...)| represents
the search space object created from calling the compiled function on some arguments.

\lstinline[emph={[2]func}, emphstyle={[2]\itshape}]|func(...).start() -> Checkpoint|

\begin{documentation}
    This begins execution of the function with the given arguments until the first branchpoint,
    i.e., a \lstinline|branchpoint()| or \lstinline|branchpoint_choose()|,
    which could be inside a nested \lstinline|searchover()| function call.
    The program state at that point is wrapped into a \lstinline|Checkpoint| object,
    which can be used to step through the function, creating checkpoints at branchpoints.
    A partial interface of \lstinline|Checkpoint| is given in \Cref{app:checkpoint}.
\end{documentation}

\lstinline[emph={[2]async_func}, emphstyle={[2]\itshape}]|async_func(...).async_start() -> AsyncCheckpoint|

\begin{documentation}
    \textit{(async method)} Async equivalent of \lstinline[emph={[2]func}, emphstyle={[2]\itshape}]|func(...).start()| for async \oursimpl{} functions.
\end{documentation}

\lstinline[emph={[2]func, search_algo, search_params}, emphstyle={[2]\itshape}]|func(...).search(search_algo: str, **search_params) -> Any|

\begin{documentation}
    This conducts search over the compiled search space using the given search algorithm and returns the final result,
    which is usually the return value (from either \lstinline|return return_value|
    or \lstinline|optional_return(return_value)|)
    from the branch with the highest latest recorded score.
    Search algorithms available in \oursimpl{} are detailed in \Cref{app:search}.
\end{documentation}

\lstinline[emph={[2]async_func, search_algo, search_params}, emphstyle={[2]\itshape}]|async_func(...).async_search(search_algo: str, **search_params) -> Any|

\begin{documentation}
    \textit{(async method)} Async equivalent of \lstinline[emph={[2]func, search_algo, search_params}, emphstyle={[2]\itshape}]|func(...).search()| for async \oursimpl{} functions.
\end{documentation}

\lstinline[emph={[2]func, search_algo, search_params}, emphstyle={[2]\itshape}]|func(...).search_multiple(search_algo: str, **search_params) -> list[tuple]|

\begin{documentation}
    This is the same as \lstinline|search()|, except it returns all results and not just the best one.
    Results are returned as a list of pairs \lstinline|(rv, score)|
    where \lstinline|rv| is the return value of a branch and \lstinline|score| is its score.
\end{documentation}

\lstinline[emph={[2]async_func, search_algo, search_params}, emphstyle={[2]\itshape}]|async_func(...).async_search_multiple(search_algo, **search_params) -> list[tuple]|

\begin{documentation}
    \textit{(Async method)} Async equivalent of \lstinline[emph={[2]func}, emphstyle={[2]\itshape}]|func(...).search_multiple()| for async \oursimpl{} functions.
\end{documentation}

\lstinline[emph={[2]func}, emphstyle={[2]\itshape}]!func.aggregate_costs: dict[str, float|int]!

\begin{documentation}
    This is a dictionary containing the aggregate costs from all \lstinline|record_cost| statements.
    Key \lstinline[emph={[2]cost_name}, emphstyle={[2]\itshape}]|"<cost_name>"| is mapped to the sum of all costs recorded with that name via
    \lstinline|record_cost(<cost_name>=...)|.
\end{documentation}

\lstinline[emph={[2]func}, emphstyle={[2]\itshape}]|func.branchpoint_step_counts: dict[Any, int]|

\begin{documentation}
    This is a dictionary that maps the name of a branchpoint to the number of times
    \lstinline|step()| has been called on a checkpoint of that branchpoint,
    over all calls to \lstinline[emph={[2]func}, emphstyle={[2]\itshape}]|func|
    since the last time \lstinline|zero_branchpoint_counts()| was called (see below).
    The dictionary will only contain step counts for named branchpoints,
    i.e., branchpoints with a \lstinline|name| parameter
    (i.e., \lstinline|branchpoint(name=...)| or \lstinline|branchpoint_choose(choices, name=...)|).
\end{documentation}

\lstinline[emph={[2]func}, emphstyle={[2]\itshape}]|func.zero_branchpoint_counts() -> None|

\begin{documentation}
    This zeros out the recorded total step counts of each named branchpoint.
\end{documentation}

\subsection{Checkpoint object interface} \label{app:checkpoint}

A \lstinline|Checkpoint| holds the program state at a branchpoint or return statement of an \oursimpl{} program's execution.

\lstinline|class Checkpoint|

\begin{documentation}
    \lstinline|step(max_protection=None, score_db_flush_queue=True) -> Checkpoint|
\end{documentation}
\begin{ddocumentation}
    This continues execution of the program starting
    from the stored program state until the next time a branchpoint is hit,
    returning a new \lstinline|Checkpoint| object.

    Any expressions protected by a
    \lstinline[emph={[2]expr, exception}, emphstyle={[2]\itshape}]|protect(expr, exception)|
    will trigger resampling whenever the exception occurs,
    up to a maximum of \lstinline|max_protection| resamplings if it is not \lstinline|None|.

    If \lstinline|score_db_flush_queue| is \lstinline|False|,
    then pending evaluations recorded through the group-evaluation
    version of \lstinline|record_score| will not be processed.

    Multiple \lstinline|step()| calls on the same \lstinline|Checkpoint| are mostly independent:
    while variable assignments are independent,
    references to variables declared as \lstinline|NoCopy|
    are shared, so that mutations to a \lstinline|NoCopy| object
    created before the current checkpoint are seen by all execution branches
    descended from this checkpoint.

    If the branchpoint is a
    \lstinline[emph={[2]choices}, emphstyle={[2]\itshape}]|branchpoint_choose(choices: Iterable)|
    instead of a regualar \lstinline|branchpoint()| statement,
    then multiple \lstinline|step()| calls iterate through
    \lstinline[emph={[2]choices}, emphstyle={[2]\itshape}]|choices|,
    and the resultant branches see the
    \lstinline[emph={[2]choices}, emphstyle={[2]\itshape}]|branchpoint_choose(choices)|
    call evaluate to the elements in
    \lstinline[emph={[2]choices}, emphstyle={[2]\itshape}]|choices|.
\end{ddocumentation}

\begin{documentation}
    \lstinline|step_sampler(max_samples=None, max_protection=None, score_db_flush_queue=True) -> Generator[Checkpoint, None, None]|
\end{documentation}
\begin{ddocumentation}
    This calls \lstinline|step()| repeatedly and yields the resultant \lstinline|Checkpoint| objects.
    This is done at most \lstinline|max_samples| is not \lstinline|None|;
    otherwise it samples forever, or until the list of choices have been exhausted
    in \lstinline|branchpoint_choose|.

    \lstinline|max_protection| specifies the \textit{total} number of resamplings
    allowed for protected expression evaluations.

    See \lstinline|Checkpoint.step()| above for \lstinline|score_db_flush_queue|.
\end{ddocumentation}

\begin{documentation}
    \lstinline|parallel_step_sampler(max_samples=None, chunk_size=None, max_protection=None, max_workers=None, score_db_flush_queue=True) -> Generator[Checkpoint, None, None]|
\end{documentation}
\begin{ddocumentation}
    Multithreaded version of \lstinline|Checkpoint.step_sampler()|,
    where \lstinline|max_workers| specifies the maximum number of threads to use
    and \lstinline|chunk_size|, if given, does parallel samplings in batches of that size.
\end{ddocumentation}

\begin{documentation}
    \lstinline|status: Status|
\end{documentation}
\begin{ddocumentation}
    The status of the checkpoint object. One of
    \lstinline|Status.RUNNING|,
    \lstinline|Status.DONE_STEPPING|,
    \lstinline|Status.RETURNED|,
    and \lstinline|Status.KILLED|.
    The \lstinline|Status.DONE_STEPPING| status
    is only possible at a \lstinline|branchpoint_choose|
    with a finite set of choices.
\end{ddocumentation}

\begin{documentation}
    \lstinline|has_return_value: bool|
\end{documentation}
\begin{ddocumentation}
    Whether there's a return value from \lstinline|return return_value| (if the checkpoint is at a return statement)
    or \lstinline|optional_return(return_value)| (if the checkpoint is at a branchpoint).
\end{ddocumentation}

\begin{documentation}
    \lstinline|return_value: Any|
\end{documentation}
\begin{ddocumentation}
    The return value of the function if it exists
    (i.e., if the checkpoint is at a return statement, or it is at a branchpoint
    following an \lstinline|optional_return| statement without an intervening branchpoint).
\end{ddocumentation}
\begin{documentation}
    \lstinline|early_stopped_search: bool|
\end{documentation}
\begin{ddocumentation}
    Whether an \lstinline|early_stopped_search()| statement has been called
    on \textit{any} branch of the program's execution.
\end{ddocumentation}
\begin{documentation}
    \lstinline!score: float|int!
\end{documentation}
\begin{ddocumentation}
    The most recent score recorded through \lstinline|record_score()|.
\end{ddocumentation}
\begin{documentation}
    \lstinline|branchpoint_params: dict|
\end{documentation}
\begin{ddocumentation}
    This is a dictionary containing the parameters of the branchpoint as specified through
    \lstinline[emph={[2]branchpoint_params}, emphstyle={[2]\itshape}]|branchpoint(**branchpoint_params)| or \lstinline[emph={[2]choices, branchpoint_params}, emphstyle={[2]\itshape}]|branchpoint_choose(choices, **branchpoint_params)|.
\end{ddocumentation}

For async \oursimpl{} functions, there's a corresponding \lstinline|AsyncCheckpoint|
with the same interface, except that certain methods are now async, and \lstinline|step_sampler()| and \lstinline|parallel_step_sampler()| have been merged into one \lstinline|async_step_sampler()|.

\lstinline|class AsyncCheckpoint|

\begin{documentation}
    \lstinline|async_step(max_protection=None, score_db_flush_queue=True) -> Checkpoint|
\end{documentation}
\begin{ddocumentation}
    \textit{(async method)} Async equivalent of \lstinline|Checkpoint.step()|.
\end{ddocumentation}

\begin{documentation}
    \lstinline|async_step_sampler(max_samples=None, chunk_size=None, max_protection=None, max_workers=None, score_db_flush_queue=True) -> AsyncGenerator[Checkpoint, None, None]|
\end{documentation}
\begin{ddocumentation}
    \textit{(async method)}
    Async equivalent of \lstinline|Checkpoint.step_sampler()|
    and \lstinline|Checkpoint.parallel_step_sampler()|.
\end{ddocumentation}

\begin{documentation}
    \lstinline|status: Status|
\end{documentation}
\begin{ddocumentation}
    See \lstinline|Checkpoint.status|.
\end{ddocumentation}

\begin{documentation}
    \lstinline|has_return_value: bool|
\end{documentation}
\begin{ddocumentation}
    See \lstinline|Checkpoint.has_return_value|.
\end{ddocumentation}

\begin{documentation}
    \lstinline|return_value: Any|
\end{documentation}
\begin{ddocumentation}
    See \lstinline|Checkpoint.return_value|.
\end{ddocumentation}

\begin{documentation}
    \lstinline|early_stopped_search: bool|
\end{documentation}
\begin{ddocumentation}
    See \lstinline|Checkpoint.early_stopped_search|.
\end{ddocumentation}
\begin{documentation}
    \lstinline!score: float|int!
\end{documentation}
\begin{ddocumentation}
    See \lstinline|Checkpoint.score|.
\end{ddocumentation}
\begin{documentation}
    \lstinline|branchpoint_params: dict|
\end{documentation}
\begin{ddocumentation}
    See \lstinline|Checkpoint.branchpoint_params|.
\end{ddocumentation}

\subsection{Search interface and search algorithms} \label{app:search}

Search algorithms are implemented over the \lstinline|Checkpoint| interface.
Parameters to a search algorithm can be specified both in the arguments
to \lstinline|search()| when invoking a compiled search space object
as well as in branchpoint parameters specified as arguments to \lstinline|branchpoint()|
and \lstinline|branchpoint_choose()| within the \oursimpl{} function.

\oursimpl{} provides several common search algorithms out-of-the-box.
The async implementations take advantage of the I/O-bound nature of LLM applications,
whereas the non-async implementations use multithreaded parallelism, which the user can disable if they wish (e.g., to prevent race conditions when there are \lstinline|NoCopy| variables).
Here is the complete list of search algorithms in the current version of \oursimpl{}:
\begin{itemize}
    \item Depth-first search (DFS)
    \item Breadth-first search (BFS)
    \item Best-first search (BeFS)
    \item Beam search
    \item Monte-Carlo tree search (MCTS), with a given value function
    \item Reexpand best-first search, a variant of BeFS where an expanded node
    can be expanded again. This was used in Case Study 3 (\Cref{sec:reflexion}).
    \item Explorative reexpand best-first search, a variant of reexpand BeFS
    where a UCB-like exploration bonus is added to the score.
\end{itemize}

The user can also implement and register their custom search algorithm
by subclassing the abstract \lstinline|Search| class.
Here, we provide a template for defining and registering a custom search algorithm:
\begin{lstlisting}[basicstyle=\ttfamily\scriptsize]
@register_search_algo(is_async=False)  # or `is_async=True` if subclassing `AsyncSearch`
class MySearch(Search):  # or `MySearch(AsyncSearch)`
    name = "my_search"
    param_names = ["param1", "param2"]  # names of branchpoint parameters that I will use

    def __init__(self, *, config1, config2, default_param1, default_param2):
        self.config1 = config1
        self.config2 = config2
        self.default_param1 = default_param1
        self.default_param2 = default_param2

    def search_generator(
        self,
        init_program_state: Checkpoint
    ) -> Generator[tuple[Any, ScoreWithCallback], None, None]:
        # or `async def async_search_generator(self, init_program_state: AsyncCheckpoint)`
        # if subclassing `AsyncSearch`
        """
        Yields pairs (return_value: Any, score_with_callback: ScoreWithCallback)
        as they are found.

        ScoreWithCallback is a wrapper around a program state's score
        - it is needed for group evaluation to work properly.
        """
        # REPLACE CODE BELOW WITH YOUR CUSTOM SEARCH ALGORITHM
        next_program_states = init_program_state.parallel_step_sampler(...)
        for next_program_state in next_program_states:
            param1 = next_program_state.get_branchpoint_param("param1", self.default_param1)
            if next_program_state.has_return_value:
                yield next_program_state.return_value, next_program_state._score_with_callback
            ...
        ...
\end{lstlisting}

\newpage

\section{The \oursimpl{} compiler} \label{app:compiler_details}

The \oursimpl{} compiler syntactically transforms an \oursimpl{} function
into an equivalent regular Python program by
conversion to continuation-passing style (CPS) and applying tail-call optimization.

For simplicity, we only describe how we compile \oursimpl{} functions
that are not async.
The compiler transformations for async \oursimpl{} functions
are nearly identical.

\subsection{CPS for branchpoints}  \label{sec:cps_branchpoints}

In this subsection, we describe how to convert a piece of code containing branchpoints
(but not any of the other EnCompass keyword primitives) into CPS.

In its simplest form, transforming a piece of code into CPS results in a function
\begin{center}
\lstinline[language=Python]|cps_function(frame: Frame, rest: Frame -> None) -> None|
\end{center}
which runs the piece of code on the variable mapping \lstinline[language=Python]|frame|
to get a new variable mapping, followed by calling the callback
\lstinline[language=Python]|rest| on that new variable mapping.
Here, the callback \lstinline[language=Python]|rest|,
called the \textit{continuation}, represents the rest of the program.

For a piece of code that doesn't contain any branchpoints, it suffices to transform
variable accesses and assignments to explicitly use \lstinline[language=Python]|frame|.
For example,
\begin{lstlisting}[basicstyle=\ttfamily\small]
x = 1
y = x + 1
\end{lstlisting}
is compiled into
\begin{lstlisting}[basicstyle=\ttfamily\small]
frame['x'] = 1
frame['y'] = frame['x'] + 1
rest(frame)
\end{lstlisting}
Note that we omit the \lstinline[language=Python]|def cps_function(frame, rest):|
in the compiled code, so technically we're compiling to the body of the CPS function.
We will call this the \textit{CPS body} to distinguish it from the \textit{CPS function}.
We defer the job of wrapping the CPS body into a function to whoever asked for the compilation.
This simplifies the issue of naming CPS functions and referring to them with the correct name.

Since the compiled CPS function explicitly runs the continuation \lstinline|rest(frame)|,
adding a branchpoint immediately after the piece of code amounts to modifying the continuation
to incorporate the search process.
So we replace \lstinline|rest(frame)| with \lstinline|branchpoint_callback(frame, rest)|,
which defines the rest of the program when we hit a branchpoint,
where \lstinline|rest| here now represents the rest of the program when we resume
from the branchpoint.
Taking the example above and adding a branchpoint at the end,
\begin{lstlisting}[basicstyle=\ttfamily\small]
x = 1
y = x + 1
branchpoint()
\end{lstlisting}
gets compiled into the following CPS body:
\begin{lstlisting}[basicstyle=\ttfamily\small]
frame['x'] = 1
frame['y'] = frame['x'] + 1
branchpoint_callback(frame, rest)
\end{lstlisting}
Here, \lstinline|branchpoint_callback(frame, rest)|
first stores the current program state
\lstinline|(frame, rest)| as a node in the search tree,
then uses the search algorithm to decide a node \lstinline|(frame1, rest1)|
in the search tree to expand, and call \lstinline|rest1(frame1.clone())|
to run the rest of the program resuming from the branchpoint
that saved the state \lstinline|(frame1, rest1)|.
Cloning \lstinline|frame1| is needed because otherwise
multiple calls to \lstinline|rest1(frame1)|
would modify the same \lstinline|frame1| object.

So far we've only defined how to transform programs with no branchpoints
and programs with one branchpoint at the end.
The transformation of a general program with branchpoints in arbitrary locations
can be defined recursively with these two base cases. For example, a program
with a branchpoint in the middle,
\begin{lstlisting}[label=lst:bp_middle, caption=Program with a branchpoint in the middle]
x = 1
branchpoint()
y = x + 1
\end{lstlisting}
is a concatenation of two programs:
\begin{lstlisting}[basicstyle=\ttfamily\small]
%$A:$%
x = 1
branchpoint()
\end{lstlisting}
and
\begin{lstlisting}[basicstyle=\ttfamily\small]
%$B:$%
y = x + 1
\end{lstlisting}
where we can apply the recursive transformation rule for concatenation,
\begin{lstlisting}[basicstyle=\ttfamily\small]
def rest(frame):
    %$CPS(B)$%
%$CPS(A)$%
\end{lstlisting}
to obtain the CPS body
\begin{lstlisting}[basicstyle=\ttfamily\small]
def rest(frame):
    frame['y'] = frame['x'] + 1
    finish_callback(frame)
frame['x'] = 1
branchpoint_callback(frame, rest)
\end{lstlisting}
Note that we have replaced \lstinline|rest(frame)| with \lstinline|finish_callback(frame)|
in the compilation of \lstinline|B| to avoid name collision with
the \lstinline|def rest(frame)|. As a result, the compiled CPS function
of the complete top-level program (AST root node) also has to reflect
this name change in its signature: \lstinline|top_level_cps_function(frame, finish_callback)|
instead of \lstinline|top_level_cps_function(frame, rest)|.
So, if \Cref{lst:bp_middle} is our entire program, then its CPS function is
\begin{lstlisting}[basicstyle=\ttfamily\small]
def top_level_cps_function(frame, finish_callback):
    def rest(frame):
        frame['y'] = frame['x'] + 1
        finish_callback(frame)
    frame['x'] = 1
    branchpoint_callback(frame, rest)
\end{lstlisting}

As a more complicated example, consider the following code:
\begin{lstlisting}[basicstyle=\ttfamily\small]
i = 0             # A
branchpoint()     # A
j = 0             # B
while i < 10:     # B
    j -= 1        # B - X
    branchpoint() # B - X
    i += 1        # B - Y
print(j)          # C
\end{lstlisting}
We've chunked up the statements at the top level into 3 pieces: $A$, $B$ and $C$.
Each chunk consists of zero or more branchpoint-free statements followed by a
statement containing a branchpoint, except for the last chunk $C$ which is branchpoint-free.
We apply the concatenation rule to $A$ and $(B; C)$, which recursively
applies the concatenation rule to $B$ and $C$.
This then recursively compiles the last statement of $B$ --- the while loop.
Compiling the while loop using the while loop rule recursively
compiles the body of the while loop using the concatenation rule on the chunks $X$ and $Y$.

The concatenation rule as applied to chunks $X$ and $Y$ gives
\begin{lstlisting}[basicstyle=\ttfamily\small]
def rest(frame):
    # CPS body of Y
    frame['i'] += 1
    continue_callback(frame)
# CPS body of X
frame['j'] -= 1
branchpoint_callback(frame, rest)
\end{lstlisting}
where to avoid name collision we replaced \lstinline|rest(frame)|
with \lstinline|continue_callback(frame)|.

Applying the while loop rule gives
\begin{lstlisting}[basicstyle=\ttfamily\small]
def body_cps_function(frame, continue_callback, break_callback):
    # CPS body of (X; Y) (from above)
    def rest(frame):
        # CPS body of Y
        frame['i'] += 1
        continue_callback(frame)
    # CPS body of X
    frame['j'] -= 1
    branchpoint_callback(frame, rest)
def while_cps_function(frame, rest):
    if frame['i'] < 10:
        body_cps_function(frame, lambda frame: while_cps_function(frame, rest))
    else:
        rest(frame)
while_cps_function(frame, rest)
\end{lstlisting}

Finally, applying the concatenation rule twice in $(A; (B; C))$ gives the CPS body
of the entire program:
\begin{lstlisting}[basicstyle=\ttfamily\small]
# CPS body of (A; B; C)
def rest(frame):
    # CPS body of (B; C)
    def rest(frame):
        # CPS body of C
        print(frame['j'])
        finish_callback(frame)
    # CPS body of B
    frame['j'] = 0
    ...  # CPS body of the while loop (from above)
# CPS body of A
frame['i'] = 0
branchpoint_callback(frame, rest)
\end{lstlisting}
And, as usual, to get the CPS function of this program, we simply wrap the above CPS body
into a \lstinline|def top_level_cps_function(frame, finish_callback)| function.

Note that the general solution to dealing with name collision is to add the correct
version of \lstinline|rest(frame)| to the end of each ``body'' in the AST during preprocessing,
so that we don't have to deal with it during conversion to CPS:
\begin{itemize}
    \item 
    At the end of the top-level program, add \lstinline|finish_callback(frame)| during preprocessing.
    During conversion to CPS, the signature of the CPS function of a top-level program
    will be \lstinline|top_level_cps_function(frame, finish_callback)| instead of \lstinline|top_level_cps_function(frame, rest)|.
    \item
    At the end of the body of a for/while loop, add \lstinline|continue_callback(frame)| during preprocessing.
    During conversion to CPS, the signature of the CPS function of the body of a for/while loop
    will be \lstinline|body_cps_function(frame, continue_callback, break_callback)| instead of \lstinline|body_cps_function(frame, rest)|. Note that this also specifies the names of the callbacks
    that \lstinline|continue| and \lstinline|break| statements in the body get converted to
    during conversion to CPS --- two birds with one stone.
    \item
    At the end of the body of an \lstinline|if| or an \lstinline|else|,
    add \lstinline|if_else_callback(frame)|.
    During conversion to CPS, the signature of the CPS function of the body of an \lstinline|if|
    will be \lstinline|if_body_cps_function(frame, if_else_callback)| instead of
    \lstinline|if_body_cps_function(frame, rest)|, and similarly for \lstinline|else|.
    \item
    At the end of the body of a function, add \lstinline|return_callback(frame.caller_frame)|.
    During conversion to CPS, the signature of the CPS function of the body of a function
    will be \lstinline|function_body_cps_function(frame, return_callback)| instead of
    \lstinline|function_body_cps_function(frame, rest)|.
\end{itemize}

We are now ready to formally write down the full set of transformations
for EnCompass programs with the simplest version of EnCompass that only has branchpoints.
For simplicity, we only describe the transformations done for synchronous code
(no async/await) where only loops, if/else statements and function definitions
have branchpoints (\texttt{with}, try-except, and \texttt{match} statements are all branchpoint-free).

\paragraph{Preprocessing} The preprocessing stage consists of the following steps:

\begin{enumerate}
    \item Convert all names $var$ to \lstinline|frame['|$var$\lstinline|']|.
    \item Add \lstinline|finish_callback(frame)| to the end of the program.
    \item Add \lstinline|continue_callback(frame)| to the end of the body of every for/while loop.
    \item Add \lstinline|if_else_callback(frame)| to the end of the body of every
    branch of every if-else statement.
    \item Add \lstinline|return_callback(frame.caller_frame)| to the end of the body of every function
    that doesn't already end in a return statement.
    \item Make the following replacements:
    \begin{itemize}
        \item \lstinline|continue| $\to$ \lstinline|continue_callback(frame)|
        \item \lstinline|break| $\to$ \lstinline|break_callback(frame)|
        \item \lstinline|return rv| $\to$ \lstinline|return_callback(frame.caller_frame, rv)|
    \end{itemize}
\end{enumerate}

\paragraph{Conversion to CPS}

Here are the general transformation rules for compiling a top-level program
or the body of a function, after the preprocessing steps described above have been completed.
\begin{enumerate}
    \item \textit{Base case --- branchpoint-free:} If $A$ has no branchpoints, make no changes.
    In other words, $CPS(A) = A$.

    \item \textit{Base case --- branchpoint:} A branchpoint
\begin{lstlisting}[basicstyle=\ttfamily\small]
branchpoint()
\end{lstlisting}
    becomes
\begin{lstlisting}[basicstyle=\ttfamily\small]
branchpoint_callback(frame, rest)
\end{lstlisting}

    \item \textit{Concatenation:} For $(A; B)$ where $A = (A'; a)$
    with $A'$ branchpoint-free and $a$ a single statement
    containing one or more branchpoints (a branchpoint or a for/while/if/else statement
    containing a branchpoint, but not e.g.\ a function definition containing a branchpoint),
\begin{lstlisting}[basicstyle=\ttfamily\small]
%$A'$%  # zero or more branchpoint-free statements
%$a$%  # single statement containing one or more branchpoints
%$B$%  # zero or more statements
\end{lstlisting}
    the compiled CPS body is
\begin{lstlisting}[basicstyle=\ttfamily\small]
def rest(frame):
    %$CPS(B)$%
%$A'$%
%$CPS(a)$%
\end{lstlisting}

    \item \textit{While loops:} For a while loop containing one or more branchpoints,
\begin{lstlisting}[basicstyle=\ttfamily\small]
while %$e$%:
    %$A$%  # contains one or more branchpoints
\end{lstlisting}
    the compiled CPS body is
\begin{lstlisting}[basicstyle=\ttfamily\small]
def body_cps_function(frame, continue_callback, break_callback):
    %$CPS(A)$%
def while_cps_function(frame, rest):
    if %$e$%:
        body_cps_function(
            frame,
            lambda frame: while_cps_function(frame, rest),
            rest
        )
    else:
        rest(frame)
while_cps_function(frame, rest)
\end{lstlisting}

    \item \textit{For loops:} For a for loop containing one or more branchpoints,
\begin{lstlisting}[basicstyle=\ttfamily\small]
for %$i$% in %$e$%:
    %$A$%  # contains one or more branchpoints
\end{lstlisting}
    the compiled CPS body is
\begin{lstlisting}[basicstyle=\ttfamily\small]
def body_cps_function(frame, continue_callback, break_callback):
    %$CPS(A)$%
def for_cps_function(frame, rest):
    try:
        %$i$% = next(frame.iterables[-1])
    except StopIteration:
        frame.iterables.pop()
        rest(frame)
        return
    def break_callback(frame):
        frame.iterables.pop()
        rest(frame)
    body_cps_function(
        frame,
        lambda frame: for_cps_function(frame, rest),
        break_callback
    )
frame.iterables.append(iter(%$e$%))
for_cps_function(frame, rest)
\end{lstlisting}

    \item \textit{If-else statements:} For an if-else statement containing
    one or more branchpoints,
\begin{lstlisting}[basicstyle=\ttfamily\small]
if %$e$%:
    %$A$%
else:
    %$B$%
\end{lstlisting}
    the compiled CPS body is
\begin{lstlisting}[basicstyle=\ttfamily\small]
def if_body_cps_function(frame, if_else_callback):
    %$CPS(A)$%
def else_body_cps_function(frame, if_else_callback):
    %$CPS(B)$%
if %$e$%:
    if_body_cps_function(frame, rest)
else:
    else_body_cps_function(frame, rest)
\end{lstlisting}
\end{enumerate}

\subsection{Tail-call optimization} \label{sec:tail-call}

There are two issues with the compiled CPS representation.
One issue is performance --- the extra function calls cause overhead,
and long for/while loops become deep recursive calls that can exceed
Python's recursion depth limit. The second issue is that defining the search
algorithm by defining \lstinline|branchpoint_callback(frame, rest)| is unnatural
and difficult. Typically, a search algorithm is implemented assuming
access to a \lstinline|step| method that returns a child of a node,
\lstinline|new_state = step(state)|.

We solve both issues via \textit{tail-call optimization}.
More specifically, every \lstinline|branchpoint_callback(frame, rest)|
is replaced with \lstinline|return frame, rest|,
and \lstinline|rest(frame)| no longer resumes from a branchpoint
to execute the rest of the program, but only executes until the next branchpoint
is hit, at which point the \lstinline|frame, rest| at that branchpoint is returned.
In other words, \lstinline|new_frame, new_rest = rest(frame.clone())|
is exactly the \lstinline|new_state = step(state)| that we need,
where we identify \lstinline|state| with \lstinline|(frame, rest)|.

With this modification, reproducing the execution of the program when all branchpoints are ignored
now involves a while loop that keeps stepping until the program finishes:
\begin{lstlisting}[basicstyle=\ttfamily\small]
frame = {}
rest = lambda frame: top_level_cps_function(frame, lambda frame: (frame, None))
while rest is not None:
    frame, rest = rest(frame)
\end{lstlisting}
And a simple DFS looks like this:
\begin{lstlisting}[basicstyle=\ttfamily\small]
frame = {}
rest = lambda frame: top_level_cps_function(frame, lambda frame: (frame, None))
stack = [(frame, rest)]
results = []
while stack:
    frame, rest = stack.pop()
    for _ in range(branching_factor):
        new_frame, new_rest = rest(frame.clone())
        if new_rest is None:
            results.append(frame)
        else:
            stack.append((new_frame, new_rest))
\end{lstlisting}
We can wrap the state \lstinline|(frame, rest)| into a \lstinline|Checkpoint| object
that provides a \lstinline|step| method wrapping \lstinline|new_frame, new_rest = rest(frame.clone())|,
and any search algorithm can now be implemented using the \lstinline|Checkpoint| interface.

We also need to modify the CPS transformation rules to return the next state
instead of running the entire continuation to completion. The details of the modifications are as follows:
\begin{enumerate}
    \item \textit{Base case --- branchpoint-free:} No change.

    \item \textit{Base case --- branchpoint:} A branchpoint
\begin{lstlisting}[basicstyle=\ttfamily\small]
branchpoint()
\end{lstlisting}
    is now compiled to
\begin{lstlisting}[basicstyle=\ttfamily\small]
return frame, rest
\end{lstlisting}

    \item \textit{Concatenation:} No change.

    \item \textit{While loops:} Prepend \lstinline|return| to these 3 lines:
\begin{lstlisting}[basicstyle=\ttfamily\small]
...
def while_cps_function(frame, rest):
    if %$e$%:
        return body_cps_function(...)   # <-
    else:
        return rest(frame)              # <-
return while_cps_function(frame, rest)  # <-
\end{lstlisting}

    \item \textit{For loops:} Prepend \lstinline|return| to these 3 lines:
\begin{lstlisting}[basicstyle=\ttfamily\small]
...
def for_cps_function(frame, rest):
    ...
        ...
        return rest(frame)            # <-
    ...
    return body_cps_function(...)     # <-
...
return for_cps_function(frame, rest)  # <-
\end{lstlisting}

    \item \textit{If-else statements:} Prepend \lstinline|return| to these 2 lines:
\begin{lstlisting}[basicstyle=\ttfamily\small]
...
if %$e$%:
    return if_body_cps_function(frame, rest)    # <-
else:
    return else_body_cps_function(frame, rest)  # <-
\end{lstlisting}
\end{enumerate}

\subsection{Other keywords}


Most other \oursimpl{} primitives provide auxiliary information,
which we store in a dictionary \lstinline|info|.
We modify our transformation rules so that \lstinline|info|
always occurs alongside \lstinline|frame|,
so the \lstinline|Checkpoint| object is now a wrapper around
the 3-tuple \lstinline|(frame, info, rest)|.
The \lstinline|Checkpoint| class implements the intended
semantics of these additional \oursimpl{} keywords
using the information stored inside \lstinline|info|
--- details which we will omit.

Note that \lstinline|info| is copied upon \lstinline|Checkpoint.step()|
similar to how \lstinline|frame| gets cloned.
In other words, stepping is now implemented as
\lstinline|new_frame, new_info, new_rest = rest(frame.clone(), info.copy())|.

For keywords that are used as standalone statements,
preprocessing is done to convert these keywords into
statements that modify \lstinline|info|.
The only exception is \lstinline|kill_branch()|,
which is transformed into a \lstinline|finish_callback()| call.
Here, we list the preprocessing transformations for all keywords
that are used as standalone statements:
\begin{itemize}
    \item \lstinline|early_stop_search()| $\to$ \lstinline|info["early_stop_search"] = True|
    \item \lstinline|kill_branch(|$e$\lstinline|)| $\to$ \lstinline|finish_callback(frame, |$e$\lstinline|, info, killed=True)|
    \item $v$\lstinline|: NeedsCopy| $\to$ \\
    \lstinline|var = |$v$\lstinline|; if var in info["nocopy"]: info["nocopy"].remove(var)|
    
    An annotated assignment is broken into two statements --- the annotation
    and the assignment --- before this transformation on the annotation occurs.
    
    \item $v$\lstinline|: NoCopy| $\to$ \lstinline|info["nocopy"].add(|$v$\lstinline|)|
    
    An annotated assignment is broken into two statements --- the annotation
    and the assignment --- before this transformation on the annotation occurs.

    \item \lstinline|optional_return(|$e$\lstinline|)| $\to$
    \lstinline|info["optional_rv"] = |$e$

    \item \lstinline|record_costs(|$keywords$\lstinline|)| $\to$
    \lstinline|info["costs"] = dict(|$keywords$\lstinline|)|

    \item \lstinline|record_score(|$args$\lstinline|)| $\to$ \\
    \lstinline|info["score"] = info["score_db"].submit_score(|$args$\lstinline|)|

    Here \lstinline|info["score_db"]| is a \lstinline|ScoreDB| object
    whose \lstinline|submit_score()| method returns a thunk that represents
    the eventual value of the score. This extra complexity is needed to implement group evaluation (\Cref{sec:self-consistency}).

    Without group evaluation,
    
    \lstinline|record_score(|$e$\lstinline|)| $\to$
    \lstinline|info["score"] = |$e$

    would suffice.
\end{itemize}

Now, the remaining keyword primitives --- 
\lstinline|branchpoint()|, \lstinline|branchpoint_choose()|, \lstinline|searchover()|
and \lstinline|protect()| --- are all used as expressions
that can be part of a larger expression or a statement.\footnote{
While \Cref{sec:cps_branchpoints} treated \lstinline|branchpoint()| as a statement,
in fact it can be used to communicate with the controller (user of the \lstinline|Checkpoint| interface),
where messages from the controller appear as the return value
of \lstinline|branchpoint()|.
}
A statement that contains one or more of these keyword primitives
needs to be partially converted to A-normal form \citep{flanagan1993anf},
where the return value
from a keyword primitive is first assigned to a temporary variable,
and its occurrence in the statement is replaced with that temporary variable.
This is done recursively for keywords nested within keywords.
For example, the statement
\begin{lstlisting}[basicstyle=\ttfamily\small]
answer = get_answer(
    branchpoint_choose([searchover(agent1(task)), protect(agent2(task), ValueError])
)
\end{lstlisting}
when converted to A-normal form will become
\begin{lstlisting}[basicstyle=\ttfamily\small]
frame.tmp_vars[0] = searchover(agent1(task))
frame.tmp_vars[1] = protect(agent2(task), ValueError)
frame.tmp_vars[2] = branchpoint_choose([frame.tmp_vars[0], frame.tmp_vars[1]])
answer = get_answer(frame.tmp_vars[2])
\end{lstlisting}
After conversion to A-normal form, each statement that assigns
the output of a keyword primitive to a temporary variable is further transformed
as follows:
\begin{itemize}
    \item \lstinline|frame.tmp_vars[|$N$\lstinline|] = branchpoint(|$kwargs$\lstinline|)| $\to$ [no change]
    \item \lstinline|frame.tmp_vars[|$N$\lstinline|] = branchpoint_choose(|$e$\lstinline|, |$kwargs$\lstinline|)| $\to$
\begin{lstlisting}[basicstyle=\ttfamily\small]
iterable = %$e$%
iterator, iterator_copy = tee(iterable)
try:
    next(iterator_copy)
except StopIteration:
    info["done_stepping"] = True
frame.tmp_vars["iterator_list"] = [iterator]
frame.tmp_vars[None] = branchpoint(%$kwargs$%)  # discard message from controller - not yet supported by branchpoint_choose()
try:
    frame.tmp_vars[%$N$%] = next(frame.tmp_vars["iterator_list"][0])
except StopIteration as e:
    raise FinishedSteppingError from e
frame.tmp_vars["iterator_list"][0], iterator_copy = tee(
    frame.tmp_vars["iterator_list"][0])
try:
    next(iterator_copy)
except StopIteration:
    info["last_branchpoint_done_stepping"] = True
\end{lstlisting}
    \item \lstinline|frame.tmp_vars[|$N$\lstinline|] = searchover(|$e$\lstinline|)| $\to$ [no change]
    \item \lstinline|frame.tmp_vars[|$N$\lstinline|] = protect(|$expr$\lstinline|, |$err$\lstinline|)| $\to$
\begin{lstlisting}[basicstyle=\ttfamily\small]
try:
    frame.tmp_vars[%$N$%] = %$expr$%
except %$err$%:
    finish_callback(frame, %$err$%, info, killed=True, protected=True)
\end{lstlisting}
\end{itemize}

Finally, note that we have to modify the CPS transformation rule
for branchpoints from \Cref{sec:tail-call},
as well as adding a CPS transformation rule for \lstinline|searchover|.
The new rules are:
\begin{itemize}
    \item \lstinline|frame.tmp_vars[|$N$\lstinline|] = branchpoint(|$kwargs$\lstinline|)|

$\to$

\begin{lstlisting}[basicstyle=\ttfamily\small]
def branchpoint_rest(frame, info, message_to_agent):
    frame.tmp_vars[%$N$%] = message_to_agent
    rest(frame, info)
return frame, info, branchpoint_rest, dict(%$kwargs$%)
\end{lstlisting}
    So now, sampling a next program state is now implemented as
    \lstinline|frame, info, branchpoint_rest, branchpoint_params = branchpoint_rest(frame.clone(), info.copy(), message_to_agent)|.

    \item \lstinline|frame.tmp_vars[|$N$\lstinline|] = searchover(|$e$\lstinline|)|

$\to$

\begin{lstlisting}[basicstyle=\ttfamily\small]
def return_rest(frame, rv, info):
    frame.tmp_vars[%$N$%] = rv
    rest(frame, info)
func_call = %$e$%
if not isinstance(func_call, SearchSpaceWithArgs):
    raise SearchoverTypeError(f"searchover(...) expects a 'SearchSpaceWithArgs' object, instead got {type(func_call)}")
func_call.compiled_cps_function(
    Frame(
        locals=func_call._args_dict,
        caller_frame=frame,
        enclosing_frame=Frame.from_closurevars(
            getclosurevars(func_call._search_space._wrapped_fn)
        )
    ),
    info,
    return_rest,
)
\end{lstlisting}
\end{itemize}










\newpage

\section{Code comparisons for case studies: base agent vs.\ \oursimpl{} agent vs.\ equivalent plain Python implementation} \label{app:plain-python}

In this appendix, for each case study, we show the code for the underlying base agent,
the agent augmented with search in \oursimpl{}, and the equivalent agent
implemented in plain Python. We annotate the changes made relative to the base agent:
\begin{itemize}
    \item \lstinline|# +n|: This line was added
    and it has $n$ words (a \textit{word} is as defined in Vim).
    \item \lstinline|# x (-m+n)|: This line was changed;
    and $m$ words were removed and $n$ words were added.
    \item \lstinline|# -m|: This line was removed and it contained $m$ words.
    \item \lstinline|# <-k|: This line (or group of omitted lines) was indent to the left by $k$ indentation levels.
    \item \lstinline|# ->k|: This line (or group of omitted lines) was indent to the right by $k$ indentation levels.
\end{itemize}
We do not count lines added that don't contain any code (i.e., that are blank or only contain a comment).

We see that, while changes made to the base agent to support search in \oursimpl{}
are minimal, significant changes are needed to support search
in the plain Python implemenation, thus demonstrating the representational advantage of \oursimpl{}.

We will omit code that remains unchanged between the base agent, the \oursimpl{} agent, and the \oursimpl{} agent's plain Python implementation.
Code segments that have been omitted are indicated by ellipses ``\lstinline|...|''.

\newpage

\subsection{Case Study 1: Code Repository Translation Agent} \label{app:python_case3}



\textit{Base agent:}

\begin{lstlisting}[basicstyle=\ttfamily\tiny, caption=Code repository translation agent, label=lst:code_trans_base]
def run_code_and_compare(method, target_code, source_code, translation_unit):
    ...  # Logging; define some variables

    if method.type == "main":

        test_inputs = None
        if "System.in" in source_code:
            # STEP 1: Write test input generation script and generate test inputs

            ...  # Prompt LLM

            ...  # Get test input format specification from LLM response
            if fatal_error:
                return 0.0

            ...  # Get test input generation script from LLM response
            if fatal_error:
                return 0.0

            ...  # Generate test inputs
            if fatal_error:
                return 0.0

        # STEP 2: Directly run codes and compare them if tested component is main function
        ...
        match = ...
        return float(match)

    # Otherwise, we have to write a main function to test the component

    ...  # Define some variables

    # STEP 1: Write test input generation script and generate test inputs

    ...  # Prompt LLM

    ...  # Get test input format specification from LLM response
    if fatal_error:
        return 0.0

    ...  # Get test input generation script from LLM response
    if fatal_error:
        return 0.0

    ...  # Generate test inputs
    if fatal_error:
        return 0.0

    # STEP 2: Run the target code with the test inputs

    ...  # Prompt LLM

    ...  # Get output format specification from LLM response
    if fatal_error:
        return 0.0

    ...  # Get target main function code from LLM response
    if fatal_error:
        return 0.0
    
    ...  # Parse target main function code
    if fatal_error:
        return 0.0

    ...  # Extract target main function AST node
    if fatal_error:
        return 0.0

    ...  # Add target main function to target code
    ...  # Run target code with main function on test inputs

    # STEP 3: Generate source code main function and run it

    ...  # Prompt LLM

    ...  # Get source main function code from LLM response
    if fatal_error:
        return 0.0

    ...  # Parse and extract source main function AST node
    if fatal_error:
        return 0.0

    ...  # Add target main function to target code
    ...  # Run target code with main function on test inputs

    matches = ...
    match_fraction = sum(matches) / len(matches)

    return match_fraction


def translate_class(translation_unit):
    ...  # Some setup (e.g., read and parse code files)
    methods_to_translate = ...
    num_methods_to_translate = len(methods_to_translate)
    translate_success_count = 0
    pass_tests_count = 0
    for method in methods_to_translate:
        target_code, translate_success = translate_method(method, target_code, source_code, translation_unit)
        if translate_success:
            translate_success_count += 1

            ...  # save target code

            if translation_unit.is_test:
                pass_tests_count += run_test_module(target_code, translation_unit)
            else:
                pass_tests_count += run_code_and_compare(
                    method,
                    target_code,
                    source_code,
                    translation_unit,
                )

    # Separately test main function (Python `if __name__ == "__main__"` block) if it's present
    if not translation_unit.is_test and ...:
        num_methods_to_translate += 1
        translate_success_count += 1
        pass_tests_count += run_code_and_compare(
            "main",
            target_code,
            source_code,
            translation_unit,
        )

    ...  # logging and saving progress

    return pass_tests_count, translate_success_count, num_methods_to_translate, new_branch


def setup_antlr4(source_code_root, target_code_root, temperature):
    source_subdir = 'src/main/antlr4'
    num_successful_translations = 0
    num_successful_parses = 0
    for root, dirs, files in os.walk(source_code_root / source_subdir):
        for file in files:
            ...  # Read antlr4 grammar file

            ...  # LLM modification if needed

            ...  # Write to target directory

            ...  # Run antlr4 to generate target Python classes

            ...  # Check if the generated files can be parsed

    return num_successful_translations + num_successful_parses


def code_translation_agent(source_code_root, target_code_root, args):
    ...  # Set up logging and git repo for saving progress

    # 0.1. Copy resource files (src/main/resources and src/test/resources)
    copy_resource_files(source_code_root, target_code_root)

    # 0.2. Set up antlr4 if applicable (src/main/antlr4)
    setup_antlr4(source_code_root, target_code_root, args.temperature)

    # 1. Get class names in topological order
    translation_units = get_translation_order_and_dependencies(source_code_root, target_code_root)

    for translation_unit in translation_units:
        # 2. Generate stubs for the class
        generate_stubs_success = generate_stubs(translation_unit)

        # 3. Translate each class
        pass_tests_count, translate_success_count, num_methods_to_translate, new_branch = translate_class(translation_unit)

        ...  # Log results

    ...  # Final logging and saving

    return final_commit


code_translation_agent(...)
\end{lstlisting}

\newpage

\textit{With \oursimpl{}:}

\begin{lstlisting}[basicstyle=\ttfamily\tiny, caption={Beam search in \oursimpl{}, 5 branchpoints excluding padding}, label=lst:code_trans_ours]
import uuid  # +2
import encompass  # +2


@encompass.compile  # +4
def branchpoint_git_commit(target_code_root, log_str="Branchpoint reached", new_branch_name="branch", **branchpoint_params):  # +17
    repo = Repo(target_code_root)  # +6
    with open(target_code_root / "commit.log", "a") as f:  # +16
        f.write(log_str + '\n')  # +9
    repo.git.add(".")  # +6
    repo.git.commit("-m", log_str)  # +10
    cur_commit = str(repo.head.commit)  # +10
    branchpoint(**branchpoint_params)  # +4
    repo.git.checkout(cur_commit)  # +8
    repo.git.switch("-c", f"{new_branch_name}-{uuid.uuid4()}")  # +16


@encompass.compile  # +4
def run_code_and_compare(method, target_code, source_code, translation_unit, base_score):  # x (-0+2)
    ...  # Logging; define some variables

    if method.type == "main":

        test_inputs = None
        if "System.in" in source_code:
            searchover(branchpoint_git_commit(  # +4
                translation_unit.target_code_root,  # +4
                f"Generate test inputs for and test {translation_unit.target_module_path}:{component_name}",  # +15
                f"bp-gen_inputs_test_main-{translation_unit.target_module_path}-{component_name}",  # +12
            ))  # +1

            # STEP 1: Write test input generation script and generate test inputs

            ...  # Prompt LLM

            ...  # Get test input format specification from LLM response
            if fatal_error:
                return 0.0

            ...  # Get test input generation script from LLM response
            if fatal_error:
                return 0.0

            ...  # Generate test inputs
            if fatal_error:
                return 0.0

        # STEP 2: Directly run codes and compare them if tested component is main function
        ...
        match = ...
        return float(match)

    # Otherwise, we have to write a main function to test the component

    ...  # Define some variables

    searchover(branchpoint_git_commit(  # +4
        translation_unit.target_code_root,  # +4
        f"Generate test inputs for {translation_unit.target_module_path}:{method}",  # +13
        f"bp-gen_inputs-{translation_unit.target_module_path}-{method}",  # +12
    ))  # +1

    # STEP 1: Write test input generation script and generate test inputs

    ...  # Prompt LLM

    ...  # Get test input format specification from LLM response
    if fatal_error:
        # pad branchpoints
        searchover(branchpoint_git_commit(translation_unit.target_code_root))  # +8
        searchover(branchpoint_git_commit(translation_unit.target_code_root))  # +8
        return 0.0

    ...  # Get test input generation script from LLM response
    if fatal_error:
        # pad branchpoints
        searchover(branchpoint_git_commit(translation_unit.target_code_root))  # +8
        searchover(branchpoint_git_commit(translation_unit.target_code_root))  # +8
        return 0.0

    ...  # Generate test inputs
    if fatal_error:
        # pad branchpoints
        searchover(branchpoint_git_commit(translation_unit.target_code_root))  # +8
        searchover(branchpoint_git_commit(translation_unit.target_code_root))  # +8
        return 0.0

    record_score(base_score + 0.01)  # +8

    searchover(branchpoint_git_commit(  # +4
        translation_unit.target_code_root,  # +4
        f"Running target code for {translation_unit.target_module_path}:{method}",  # +13
        f"bp-run_target-{translation_unit.target_module_path}-{method}",  # +12
    ))  # +1

    # STEP 2: Run the target code with the test inputs

    ...  # Prompt LLM

    ...  # Get output format specification from LLM response
    if fatal_error:
        # pad branchpoints
        searchover(branchpoint_git_commit(translation_unit.target_code_root))  # +8
        return 0.0

    ...  # Get target main function code from LLM response
    if fatal_error:
        # pad branchpoints
        searchover(branchpoint_git_commit(translation_unit.target_code_root))  # +8
        return 0.0
    
    ...  # Parse target main function code
    if fatal_error:
        # pad branchpoints
        searchover(branchpoint_git_commit(translation_unit.target_code_root))  # +8
        return 0.0

    ...  # Extract target main function AST node
    if fatal_error:
        # pad branchpoints
        searchover(branchpoint_git_commit(translation_unit.target_code_root))  # +8
        return 0.0

    ...  # Add target main function to target code
    ...  # Run target code with main function on test inputs

    record_score(base_score + 0.02)  # +8

    searchover(branchpoint_git_commit(  # +4
        translation_unit.target_code_root,  # +4
        f"Running source code for {translation_unit.target_module_path}:{method}",  # +13
        f"bp-run_source-{translation_unit.target_module_path}-{method}",  # +12
    ))  # +1

    # STEP 3: Generate source code main function and run it

    ...  # Prompt LLM

    ...  # Get source main function code from LLM response
    if fatal_error:
        return 0.0

    ...  # Parse and extract source main function AST node
    if fatal_error:
        return 0.0

    ...  # Add target main function to target code
    ...  # Run target code with main function on test inputs

    matches = ...
    match_fraction = sum(matches) / len(matches)

    return match_fraction


@encompass.compile  # +4
def translate_class(translation_unit):
    ...  # Some setup (e.g., read and parse code files)
    methods_to_translate = ...
    num_methods_to_translate = len(methods_to_translate)
    translate_success_count = 0
    pass_tests_count = 0
    for method in methods_to_translate:
        searchover(branchpoint_git_commit(  # +4
            translation_unit.target_code_root,  # +4
            f"Begin {translation_unit.source_class_path} translation of {method}.",  # +13
            f"bp-translate-{translation_unit.source_class_path}-{method}",  # +12
        ))  # +1

        target_code, translate_success = translate_method(method, target_code, source_code, translation_unit)
        if translate_success:
            translate_success_count += 1
            record_score(translate_success_count + pass_tests_count)  # +6

            ...  # save target code

            if translation_unit.is_test:
                pass_tests = run_test_module(target_code, translation_unit)
                pass_tests_count += pass_tests
            else:
                pass_tests_count += searchover(run_code_and_compare(  # x (-0+2)
                    method,
                    target_code,
                    source_code,
                    translation_unit,
                    base_score = translate_success_count + pass_tests_count  # +5
                ))  # x (-0+1)
            record_score(translate_success_count + pass_tests_count)  # +6

    # Separately test main function (Python `if __name__ == "__main__"` block) if it's present
    if not translation_unit.is_test and ...:
        num_components_to_translate += 1
        translate_success_count += 1
        pass_tests_count += searchover(run_code_and_compare(  # x (-0+2)
            "main",
            target_code,
            source_code,
            translation_unit,
            base_score = translate_success_count + pass_tests_count  # +5
        ))  # x (-0+1)
        record_score(translate_success_count + pass_tests_count)  # +6

    ...  # logging and saving progress

    return pass_tests_count, translate_success_count, len(methods_to_translate), new_branch


@encompass.compile  # +4
def setup_antlr4(source_code_root, target_code_root, temperature):
    source_subdir = 'src/main/antlr4'
    num_successful_translations = 0
    num_successful_parses = 0
    for root, dirs, files in os.walk(source_code_root / source_subdir):
        for file in files:
            ...  # Read antlr4 grammar file

            searchover(branchpoint_git_commit(  # +4
                target_code_root,  # +2
                f"Translate antlr4 grammar {source_file_path.stem}",  # +10
                f"bp-translate_antlr4_grammar-{source_file_path.stem}",  # +10
            ))  # +1

            ...  # LLM modification if needed

            ...  # Write to target directory

            ...  # Run antlr4 to generate target Python classes

            ...  # Check if the generated files can be parsed

            record_score(num_successful_translations + num_successful_parses)  # +6

    return num_successful_translations + num_successful_parses


@encompass.compile  # +4
def code_translation_agent(source_code_root, target_code_root, args):
    ...  # Set up logging and git repo for saving progress

    # 0.1. Copy resource files (src/main/resources and src/test/resources)
    copy_resource_files(source_code_root, target_code_root)

    # 0.2. Set up antlr4 if applicable (src/main/antlr4)
    total_score = searchover(setup_antlr4(source_code_root, target_code_root, args.temperature))  # x (-0+5)

    # 1. Get class names in topological order
    translation_units = get_translation_order_and_dependencies(source_code_root, target_code_root)

    for translation_unit in translation_units:
        searchover(branchpoint_git_commit(  # +4
            target_code_root,  # +2
            f"Begin {translation_unit.source_class_path} translation.",  # +10
            f"bp-translate-{translation_unit.source_class_path}",  # +10
        ))  # +1

        # 2. Generate stubs for the class
        generate_stubs_success = generate_stubs(translation_unit)

        total_score += generate_stubs_success  # +3
        record_score(total_score)  # +4

        # 3. Translate each class
        searchover(branchpoint_git_commit(translation_unit.target_code_root, branching=1))  # +12
        translate_class_results = translate_class(translation_unit).search_multiple("beam", beam_width=2, default_branching=2)  # x (-0+14)
        (pass_tests_count, translate_success_count, num_methods_to_translate, new_branch), _ = branchpoint_choose(translate_class_results, branching=len(translate_class_results))  # x (see above)

        # "+1" to prevent agent from "cheating" (have very few e.g. zero stubs to implement)
        total_score += pass_tests_count / (num_methods_to_translate + 1)  # +9
        record_score(total_score)  # +4

        ...  # Log results

    ...  # Final logging and saving

    return final_commit


code_translation_agent(...).search("beam", beam_width=3, default_branching=3)  # x (-0+13)
\end{lstlisting}

\newpage

\textit{Without \oursimpl{}:}
Explicitly defining a state machine to support general search not only significantly obscures the underlying agent logic, but is also prone to bugs such as \texttt{KeyError} when accessing the dictionary \texttt{cur\_state} that stores all the variables. A lot of newly added code is for bookkeeping to maintain a persistent state, which is implemented as a dictionary that stores the variables of the base agent.

\begin{lstlisting}[basicstyle=\ttfamily\tiny, caption={Beam search implemented in plain Python}, label=lst:code_trans_plain]
import uuid  # +2
import numpy as np  # +4


def git_commit(target_code_root, log_str="Branchpoint reached"):  # +10
    repo = Repo(target_code_root)  # +6
    with open(target_code_root / "commit.log", "a") as f:  # +16
        f.write(log_str + '\n')  # +9
    repo.git.add(".")  # +6
    repo.git.commit("-m", log_str)  # +10
    cur_commit = str(repo.head.commit)  # +10
    return cur_commit  # +2


def checkout_new_branch(target_code_root, cur_commit, new_branch_name="branch"):  # +11
    repo = Repo(target_code_root)  # +6
    repo.git.checkout(cur_commit)  # +8
    repo.git.switch("-c", f"{new_branch_name}-{uuid.uuid4()}")  # +16


def run_code_and_compare_prelude(cur_state, cur_commit, cur_score):  # x (-8+6)
    # Get used variables from `cur_state`
    method = cur_state["method"]  # +6
    target_code = cur_state["target_code"]  # +6
    source_code = cur_state["source_code"]  # +6
    translation_unit = cur_state["translation_unit"]  # +6

    ...  # Logging; define some variables

    if method.type == "main":

        test_inputs = None

        # Store new variables to `new_state`
        new_state = cur_state.copy()  # +6
        new_state["method"] = method  # +6
        new_state["run_code_log_path"] = run_code_log_path  # +6
        new_state["test_inputs"] = test_inputs  # +6
        new_state["dependency_files_str"] = dependency_files_str  # +6
        new_state["fatal_error"] = fatal_error  # +6

        if "System.in" in source_code:
            # git commit
            commit = git_commit(  # +4
                translation_unit.target_code_root,  # +4
                f"Generate test inputs for and test {translation_unit.target_module_path}:{method}",  # +15
            )  # +1

            return new_state, run_code_and_compare_gen_test_inputs_existing_main, cur_score, commit  # +8

        return run_code_and_compare_run_target_source_codes_existing_main(new_state, cur_score, cur_commit)  # +9

    ...  # Define some variables

    # Store newly defined variables to `new_state`
    new_state = cur_state.copy()  # +6
    new_state["method"] = method  # +6
    new_state["run_code_log_path"] = run_code_log_path  # +6
    new_state["num_test_inputs"] = num_test_inputs  # +6
    new_state["target_code_without_main"] = target_code_without_main  # +6
    new_state["target_code_with_dummy_main"] = target_code_with_dummy_main  # +6
    new_state["dependency_files_str"] = dependency_files_str  # +6
    new_state["fatal_error"] = fatal_error  # +6

    # git commit
    commit = git_commit(  # +4
        translation_unit.target_code_root,  # +4
        f"Generate test inputs for {translation_unit.target_module_path}:{method}",  # +13
    )  # +1
    return new_state, run_code_and_compare_gen_test_inputs, cur_score, commit  # +8


def run_code_and_compare_gen_test_inputs_existing_main(cur_state, cur_commit, cur_score):  # +9
    # Get used variables from `cur_state`
    method = cur_state["method"]  # +6
    translation_unit = cur_state["translation_unit"]  # +6
    run_code_log_path = cur_state["run_code_log_path"]  # +6
    dependency_files_str = cur_state["dependency_files_str"]  # +6
    fatal_error = cur_state["fatal_error"]  # +6

    checkout_new_branch(  # +2
        cur_commit,  # +2
        f"bp-gen_inputs_test_main-{translation_unit.target_module_path}-{method}",  # +12
    )  # +1

    # Prepare for next step
    new_state = cur_state.copy()  # +6

    # Write test input generation script and generate test inputs

    ...  # Prompt LLM  # <-2

    ...  # Get test input format specification from LLM response  # <-2
    if fatal_error:  # <-2
        return new_state, translate_class_postlude_2, cur_score, None  # <-2  # x (-3+7)

    ...  # Get test input generation script from LLM response  # <-2
    if fatal_error:  # <-2
        return new_state, translate_class_postlude_2, cur_score, None  # <-2  # x (-3+7)

    ...  # Generate test inputs  # <-2
    if fatal_error:  # <-2
        return new_state, translate_class_postlude_2, cur_score, None  # <-2  # x (-3+7)

    # Store newly defined variables to `new_state`
    new_state["stdin_format"] = stdin_format  # +6
    new_state["gen_inputs_code"] = gen_inputs_code  # +6
    new_state["test_inputs"] = test_inputs  # +6
    new_state["fatal_error"] = fatal_error  # +6

    return run_code_and_compare_run_target_source_codes_existing_main(new_state, cur_score, cur_commit)  # +9


def run_code_and_compare_run_target_source_codes_existing_main(cur_state, cur_commit, cur_score):  # +9
    # Get used variables from `cur_state`
    method = cur_state["method"]  # +6
    translation_unit = cur_state["translation_unit"]  # +6
    run_code_log_path = cur_state["run_code_log_path"]  # +6
    dependency_files_str = cur_state["dependency_files_str"]  # +6
    stdin_format = cur_state["stdin_format"]  # +6
    test_inputs = cur_state["test_inputs"]  # +6
    fatal_error = cur_state["fatal_error"]  # +6

    # Directly run codes and compare them if tested method is main function
    ...  # <-1
    match = ...  # <-1

    # Store new variables to `new_state`
    new_state = cur_state.copy()  # +6
    new_state["pass_tests_count"] += float(match)  # +9

    # Compute new score; decide next step
    score = new_state["translate_success_count"] + new_state["pass_tests_count"]  # +11
    return new_state, translate_class_postlude_2, score, None  # <-2  # x (-3+7)


def run_code_and_compare_gen_test_inputs(cur_state, cur_commit, cur_score):  # +9
    # Get used variables from `cur_state`
    method = cur_state["method"]  # +6
    translation_unit = cur_state["translation_unit"]  # +6
    run_code_log_path = cur_state["run_code_log_path"]  # +6
    num_test_inputs = cur_state["num_test_inputs"]  # +6
    target_code_without_main = cur_state["target_code_without_main"]  # +6
    dependency_files_str = cur_state["dependency_files_str"]  # +6
    fatal_error = cur_state["fatal_error"]  # +6

    checkout_new_branch(  # +2
        cur_commit,  # +2
        f"bp-gen_inputs-{translation_unit.target_module_path}-{method}",  # +12
    )  # +1

    # STEP 1: Write test input generation script and generate test inputs

    ...  # Prompt LLM

    ...  # Get test input format specification from LLM response
    if fatal_error:
        commit = git_commit(translation_unit.target_code_root)  # +8
        return cur_state, run_code_and_compare_idle_1, cur_score, commit  # x (-3+7)

    ...  # Get test input generation script from LLM response
    if fatal_error:
        commit = git_commit(translation_unit.target_code_root)  # +8
        return cur_state, run_code_and_compare_idle_1, cur_score, commit  # x (-3+7)

    ...  # Generate test inputs
    if fatal_error:
        commit = git_commit(translation_unit.target_code_root)  # +8
        return cur_state, run_code_and_compare_idle_1, cur_score, commit  # x (-3+7)

    # Store newly defined variables to `new_state`
    new_state = cur_state.copy()  # +6
    new_state["stdin_format"] = stdin_format  # +6
    new_state["gen_inputs_code"] = gen_inputs_code  # +6
    new_state["test_inputs_list"] = test_inputs_list  # +6
    new_state["fatal_error"] = fatal_error  # +6

    # Compute score and git commit
    score = new_state["base_score"] + 0.01  # +10
    commit = git_commit(  # +4
        translation_unit.target_code_root,  # +4
        f"Running target code for {translation_unit.target_module_path}:{method}",  # +13
    )  # +1
    return new_state, run_code_and_compare_run_target_code, score, commit  # +8


def run_code_and_compare_idle_1(cur_state, cur_commit, cur_score):  # +9
    translation_unit = cur_state["translation_unit"]  # +6
    checkout_new_branch(cur_commit)  # +4
    commit = git_commit(translation_unit.target_code_root)  # +8
    return cur_state, run_code_and_compare_idle_2, cur_score, commit  # +8


def run_code_and_compare_run_target_code(cur_state, cur_commit, cur_score):  # +9
    # Get used variables from `cur_state`
    method = cur_state["method"]  # +6
    translation_unit = cur_state["translation_unit"]  # +6
    run_code_log_path = cur_state["run_code_log_path"]  # +6
    target_code_with_dummy_main = cur_state["target_code_with_dummy_main"]  # +6
    dependency_files_str = cur_state["dependency_files_str"]  # +6
    stdin_format = cur_state["stdin_format"]  # +6
    gen_inputs_code = cur_state["gen_inputs_code"]  # +6
    test_inputs_list = cur_state["test_inputs_list"]  # +6
    fatal_error = cur_state["fatal_error"]  # +6

    checkout_new_branch(  # +2
        cur_commit,  # +2
        f"bp-run_target-{translation_unit.target_module_path}-{method}",  # +12
    )  # +1

    # Run the target code with the test inputs

    ...  # Prompt LLM

    ...  # Get output format specification from LLM response
    if fatal_error:
        commit = git_commit(translation_unit.target_code_root)  # +8
        return cur_state, run_code_and_compare_idle_2, cur_score, commit  # x (-3+7)

    ...  # Get target main function code from LLM response
    if fatal_error:
        commit = git_commit(translation_unit.target_code_root)  # +8
        return cur_state, run_code_and_compare_idle_2, cur_score, commit  # x (-3+7)
    
    ...  # Parse target main function code
    if fatal_error:
        commit = git_commit(translation_unit.target_code_root)  # +8
        return cur_state, run_code_and_compare_idle_2, cur_score, commit  # x (-3+7)

    ...  # Extract target main function AST node
    if fatal_error:
        commit = git_commit(translation_unit.target_code_root)  # +8
        return cur_state, run_code_and_compare_idle_2, cur_score, commit  # x (-3+7)

    ...  # Add target main function to target code
    ...  # Run target code with main function on test inputs

    # Store newly defined variables to `new_state`
    new_state = cur_state.copy()  # +6
    new_state["stdout_format"] = stdout_format  # +6
    new_state["target_code_with_main"] = target_code_with_main  # +6
    new_state["run_target_results"] = run_target_results  # +6
    new_state["fatal_error"] = fatal_error  # +6

    # Compute score and git commit
    score = new_state["base_score"] + 0.02  # +10
    commit = git_commit(  # +4
        translation_unit.target_code_root,  # +4
        f"Running source code for {translation_unit.target_module_path}:{method}",  # +13
    )  # +1
    return new_state, run_code_and_compare_run_source_code, score, commit  # +8


def run_code_and_compare_idle_2(cur_state, cur_commit, cur_score):  # +9
    translation_unit = cur_state["translation_unit"]  # +6
    checkout_new_branch(cur_commit)  # +4

    # Store new variables to `new_state`
    new_state = cur_state.copy()  # +6
    new_state["method_idx"] += 1  # +6

    # git commit; decide next step
    if new_state["method_idx"] == len(new_state["methods_to_translate"]):  # +12
        commit = None  # +3
        next_step = translate_class_postlude_1  # +3
    else:  # +2
        new_state["method"] = new_state["methods_to_translate"][new_state["method_idx"]]  # +13
        commit = git_commit(  # +4
            translation_unit.target_code_root,  # +4
            f"Begin {translation_unit.source_class_path} translation of {new_state["method"]}.",  # +15
        )  # +1
        next_step = translate_method_and_save  # +3
    return new_state, next_step, cur_score, commit  # +8


def run_code_and_compare_run_source_code(cur_state, cur_commit, cur_score):  # +9
    # Get used variables from `cur_state`
    method = cur_state["method"]  # +6
    translation_unit = cur_state["translation_unit"]  # +6
    run_code_log_path = cur_state["run_code_log_path"]  # +6
    source_code = cur_state["source_code"]  # +6
    target_code_with_main = cur_state["target_code_with_main"]  # +6
    stdin_format = cur_state["stdin_format"]  # +6
    stdout_format = cur_state["stdout_format"]  # +6
    test_inputs_list = cur_state["test_inputs_list"]  # +6
    fatal_error = cur_state["fatal_error"]  # +6

    checkout_new_branch(  # +2
        cur_commit,  # +2
        f"bp-run_source-{translation_unit.target_module_path}-{method}",  # +12
    )  # +1

    # Prepare for next step
    new_state = cur_state.copy()  # +6
    new_state["method_idx"] += 1  # +6

    # Generate source code main function and run it

    ...  # Prompt LLM

    ...  # Get source main function code from LLM response
    if fatal_error:
        # git commit; decide next step
        if new_state["method_idx"] == len(new_state["methods_to_translate"]):  # +12
            commit = None  # +3
            next_step = translate_class_postlude_1  # +3
        else:  # +2
            new_state["method"] = new_state["methods_to_translate"][new_state["method_idx"]]  # +13
            commit = git_commit(  # +4
                translation_unit.target_code_root,  # +4
                f"Begin {translation_unit.source_class_path} translation of {new_state["method"]}.",  # +15
            )  # +1
            next_step = translate_method_and_save  # +3
        return new_state, next_step, cur_score, commit  # x (-3+7)

    ...  # Parse and extract source main function AST node
    if fatal_error:
        # git commit; decide next step
        if new_state["method_idx"] == len(new_state["methods_to_translate"]):  # +12
            commit = None  # +3
            next_step = translate_class_postlude_1  # +3
        else:  # +2
            new_state["method"] = new_state["methods_to_translate"][new_state["method_idx"]]  # +13
            commit = git_commit(  # +4
                translation_unit.target_code_root,  # +4
                f"Begin {translation_unit.source_class_path} translation of {new_state["method"]}.",  # +15
            )  # +1
            next_step = translate_method_and_save  # +3
        return new_state, next_step, cur_score, commit  # x (-3+7)

    ...  # Add target main function to target code
    ...  # Run target code with main function on test inputs

    matches = ...
    match_fraction = sum(matches) / len(matches)

    # Store new variables to `new_state`
    new_state = cur_state.copy()  # +6
    new_state["pass_tests_count"] += match_fraction  # +6

    # Compute new score; git commit; decide next step
    score = new_state["translate_success_count"] + new_state["pass_tests_count"]  # +11
    if new_state["method_idx"] == len(new_state["methods_to_translate"]):  # +12
        commit = None  # +3
        next_step = translate_class_postlude_1  # +3
    else:  # +2
        new_state["method"] = new_state["methods_to_translate"][new_state["method_idx"]]  # +13
        commit = git_commit(  # +4
            translation_unit.target_code_root,  # +4
            f"Begin {translation_unit.source_class_path} translation of {new_state["method"]}.",  # +15
        )  # +1
        next_step = translate_method_and_save  # +3
    return new_state, next_step, score, commit  # x (-1+7)


def translate_class_prelude(cur_state, cur_commit, cur_score):  # +9
    # Get used variables from `cur_state`
    translation_unit = cur_state["translation_unit"]  # +6

    ...  # Some setup (e.g., read and parse code files)
    methods_to_translate = ...
    num_methods_to_translate = len(methods_to_translate)
    translate_success_count = 0
    pass_tests_count = 0
    # for method in methods_to_translate:  # -5

    # Store newly defined variables to `new_state`
    new_state = cur_state.copy()  # +6
    new_state["methods_to_translate"] = methods_to_translate  # +6
    new_state["num_methods_to_translate"] = num_methods_to_translate  # +6
    new_state["translate_success_count"] = translate_success_count  # +6
    new_state["pass_tests_count"] = pass_tests_count  # +6
    new_state["method_idx"] = 0  # +6
    new_state["method"] = methods_to_translate[0]  # +9

    # git commit
    commit = git_commit(  # +4
        translation_unit.target_code_root,  # +4
        f"Begin {translation_unit.source_class_path} translation of {new_state["method"]}.",  # +15
    )  # +1
    return new_state, translate_method_and_save, cur_score, commit  # +8


def translate_method_and_save(cur_state, cur_commit, cur_score):  # +9
    # Get used variables from `cur_state`
    method = cur_state["method"]  # +6
    translation_unit = cur_state["translation_unit"]  # +6
    source_code = cur_state["source_code"]  # +6
    target_code = cur_state["target_code"]  # +6
    translate_success_count = cur_state["translate_success_count"]  # +6
    pass_tests_count = cur_state["pass_tests_count"]  # +6

    checkout_new_branch(  # +2
        cur_commit,  # +2
        f"bp-translate-{translation_unit.source_class_path}-{method}",  # +12
    )  # +1

    new_state = cur_state.copy()  # +6

    target_code, translate_success = translate_method(method, target_code, source_code, translation_unit)  # <-1
    if translate_success:  # <-1
        translate_success_count += 1  # <-1
        score = translate_success_count + pass_tests_count  # +5

        ...  # save target code  # <-1

        if translation_unit.is_test:  # <-1
            pass_tests_count += run_test_module(target_code, translation_unit)  # <-1
        else:  # <-1
            # pass_tests_count += run_code_and_compare(  # -4
            #     method,  # -2
            #     target_code,  # -2
            #     source_code,  # -2
            #     translation_unit,  # -2
            # )  # -1

            # Store newly defined variables to `new_state`
            new_state["base_score"] = translate_success_count + pass_tests_count  # +8

            return run_code_and_compare_prelude(new_state, cur_score, cur_commit)  # +9

    # Store new variables to `new_state`
    new_state["method_idx"] += 1  # +6

    # git commit; decide next step
    if new_state["method_idx"] == len(new_state["methods_to_translate"]):  # +12
        commit = None  # +3
        next_step = translate_class_postlude_1  # +3
    else:  # +2
        new_state["method"] = new_state["methods_to_translate"][new_state["method_idx"]]  # +13
        commit = git_commit(  # +4
            translation_unit.target_code_root,  # +4
            f"Begin {translation_unit.source_class_path} translation of {new_state["method"]}.",  # +15
        )  # +1
        next_step = translate_method_and_save  # +3
    return new_state, next_step, cur_score, commit  # +8


def translate_class_postlude_1(cur_state, cur_commit, cur_score):  # +9
    # Get used variables from `cur_state`
    translation_unit = cur_state["translation_unit"]  # +6
    target_code = cur_state["target_code"]  # +6
    translate_success_count = cur_state["translate_success_count"]  # +6
    pass_tests_count = cur_state["pass_tests_count"]  # +6
    num_methods_to_translate = cur_state["num_methods_to_translate"]  # +6

    new_state = state.copy()  # +6

    # Separately test main function (Python `if __name__ == "__main__"` block) if it's present
    if not translation_unit.is_test and ...:
        num_methods_to_translate += 1
        translate_success_count += 1

        # Store newly defined variables to `new_state`
        new_state["num_methods_to_translate"] = num_methods_to_translate  # +6
        new_state["translate_success_count"] = translate_success_count  # +6
        new_state["method"] = "main"  # +8
        new_state["base_score"] = translate_success_count + pass_tests_count  # +8

        return run_code_and_compare_prelude(new_state, cur_score, cur_commit)  # +9
    
    return translate_class_postlude_2(new_state, cur_commit, cur_score)  # +9


def translate_class_postlude_2(cur_state, cur_commit, cur_score):  # +9
    # Get used variables from `cur_state`
    translation_unit = cur_state["translation_unit"]  # +6
    translate_success_count = cur_state["translate_success_count"]  # +6
    pass_tests_count = cur_state["pass_tests_count"]  # +6
    num_methods_to_translate = cur_state["num_methods_to_translate"]  # +6

    ...  # logging and saving progress

    return_value = (pass_tests_count, translate_success_count, num_methods_to_translate, new_branch)  # x (see below)
    return return_value, None, cur_score, None  # x (-0+11)


def translate_class(translation_unit, beam_width, branching):  # x (-0+4)
    # Use beam search to translate a class method-by-method

    init_state = {"translation_unit": translation_unit}  # +7
    init_step = translate_class_prelude  # +3
    init_commit = None  # +3
    init_score = 0.0  # +5

    beam = [init_step(init_state, init_commit, init_score)]  # +11
    results = []  # +3
    while len(beam) > 0:  # +8
        new_program_states_list = []  # +3
        for state, step, score, commit in beam:  # +11
            new_program_states = [step(state, commit, score) for _ in range(branching)]  # +18
            new_program_states.sort(key=lambda x: x[2], reverse=True)  # +17
            new_program_states_list.append(new_program_states)  # +6
        not_done_new_program_states = []  # +3
        for i in range(len(new_program_states_list[0])):  # +11
            # random permutation of indices to break ties
            for j in np.random.permutation(len(new_program_states_list)):  # +13
                new_program_state = new_program_states_list[j][i]  # +8
                new_state, new_step, new_score, new_commit = new_program_state  # +9
                if new_step is None:  # +5
                    results.append((new_state, new_score))  # +8
                else:  # +2
                    not_done_new_program_states.append(new_program_state)  # +6
        not_done_new_program_states.sort(  # +4
            key=lambda program_state: program_state.score, reverse=True  # +12
        )  # +1
        beam = not_done_new_program_states[:beam_width]  # +6
    return results  # +2


def setup_antlr4_prelude(cur_state, cur_commit, cur_score):  # x (-6+6)
    # Get used variables from `cur_state`
    source_code_root = cur_state["source_code_root"]  # +6

    new_state = cur_state.copy()  # +6

    source_subdir = 'src/main/antlr4'
    new_state["num_successful_translations"] = 0  # +3
    new_state["num_successful_parses"] = 0  # +3
    new_state["root_dirs_files_list"] = list(os.walk(source_code_root / source_subdir))  # x (-8+8)
        # for file in files:  # -5

    new_state["root_dirs_files_idx"] = 0  # +6
    new_state["file_idx"] = 0  # +6

    root, dirs, files = new_state["root_dirs_files_list"][new_state["root_dirs_files_idx"]]  # +14
    file = files[new_state["file_idx"]]  # +8

    ...  # Read antlr4 grammar file  # <-2

    # Store newly defined variables to `new_state`
    new_state["source_file_path"] = source_file_path  # +6
    new_state["grammar_content"] = grammar_content  # +6

    # git commit
    commit = git_commit(  # +4
        target_code_root,  # +2
        f"Translate antlr4 grammar {source_file_path.stem}",  # +10
    )  # +1

    return new_state, setup_antlr4_body, cur_score, commit  # +8


def setup_antlr4_body(cur_state, cur_commit, cur_score):  # +9
    # Get used variables from `cur_state`
    source_code_root = cur_state["source_code_root"]  # +6
    target_code_root = cur_state["target_code_root"]  # +6
    temperature = cur_state["temperature"]  # +6
    num_successful_translations = cur_state["num_successful_translations"]  # +6
    num_successful_parses = cur_state["num_successful_parses"]  # +6
    root, dirs, files = cur_state["root_dirs_files_list"][cur_state["root_dirs_files_idx"]]  # +14
    file = files[cur_state["file_idx"]]  # +8
    source_file_path = cur_state["source_file_path"]  # +6
    grammar_content = cur_state["grammar_content"]  # +6

    checkout_new_branch(  # +2
        cur_commit,  # +2
        f"bp-translate_antlr4_grammar-{source_file_path.stem}",  # +10
    )  # +1

    ...  # LLM modification if needed  # <-2

    ...  # Write to target directory  # <-2

    ...  # Run antlr4 to generate target Python classes  # <-2

    ...  # Check if the generated files can be parsed  # <-2

    cur_score = num_successful_translations + num_successful_parses  # +5

    new_state = cur_state.copy()  # +6

    # Increment to next loop iteration
    new_state["root_dirs_files_idx"] += 1  # +6
    new_state["file_idx"] += 1  # +6
    if new_state["file_idx"] == len(files):  # +10
        # Inner for loop completed -- increment outer for loop index
        new_state["root_dirs_files_idx"] += 1  # +6
        new_state["file_idx"] = 0  # +6
        if new_state["root_dirs_files_idx"] == len(new_state["root_dirs_files_list"]):  # +12
            # Outer for loop completed -- return to code repo translation agent
            new_state["total_score"] = num_successful_translations + num_successful_parses  # x (see below)
            return code_translation_agent_prelude_2(new_state, cur_commit, cur_score)  # x (-0+13)

    root, dirs, files = new_state["root_dirs_files_list"][new_state["root_dirs_files_idx"]]  # +14
    file = files[new_state["file_idx"]]  # +8

    ...  # Read antlr4 grammar file  # <-2

    # Store newly defined variables to `new_state`
    new_state["source_file_path"] = source_file_path  # +6
    new_state["grammar_content"] = grammar_content  # +6

    # git commit
    commit = git_commit(  # +4
        target_code_root,  # +2
        f"Translate antlr4 grammar {source_file_path.stem}",  # +10
    )  # +1

    return new_state, setup_antlr4_body, cur_score, commit  # +8


def code_translation_agent_prelude_1(cur_state, cur_commit, cur_score):  # +9
    # Get used variables from `cur_state`
    source_code_root = cur_state["source_code_root"]  # +6
    target_code_root = cur_state["target_code_root"]  # +6

    ...  # Set up logging and git repo for saving progress

    # 0.1. Copy resource files (src/main/resources and src/test/resources)
    copy_resource_files(source_code_root, target_code_root)

    # Store newly defined variables to `new_state`
    new_state = cur_state.copy()  # +6
    new_state["repo"] = repo  # +6
    new_state["results"] = results  # +6
    new_state["temperature"] = cur_state["args"].temperature  # +10

    return setup_antlr4_prelude(new_state, cur_commit, cur_score)


def code_translation_agent_prelude_2(cur_state, cur_commit, cur_score):  # +9
    # Get used variables from `cur_state`
    source_code_root = cur_state["source_code_root"]  # +6
    target_code_root = cur_state["target_code_root"]  # +6

    # Get class names in topological order
    translation_units = get_translation_order_and_dependencies(source_code_root, target_code_root)

    # Store newly defined variables to `new_state`
    new_state = cur_state.copy()  # +6
    new_state["translation_units"] = translation_units  # +6
    new_state["translation_unit_idx"] = 0  # +6

    # git commit
    translation_unit = new_state["translation_units"][new_state["translation_unit_idx"]]  # +10
    commit = git_commit(  # +4
        translation_unit.target_code_root,  # +4
        f"Begin {translation_unit.source_class_path} translation.",  # +10
    )  # +1
    return new_state, code_translation_agent_generate_stubs, cur_score, commit  # +8


def code_translation_agent_generate_stubs(cur_state, cur_commit, cur_score):  # +9
    # Get used variables from `cur_state`
    translation_unit = cur_state["translation_units"][cur_state["translation_unit_idx"]]  # +10
    total_score = cur_state["total_score"]  # +6

    checkout_new_branch(  # +2
        cur_commit,  # +2
        f"bp-translate-{translation_unit.source_class_path}",  # +10
    )  # +1

    # Generate stubs for the class
    generate_stubs_success = generate_stubs(translation_unit)  # <-1

    total_score += generate_stubs_success  # +3

    # Store newly defined variables to `new_state`
    new_state = cur_state.copy()  # +6
    new_state["generate_stubs_success"] = generate_stubs_success  # +6
    new_state["total_score"] = total_score  # +6

    # git commit
    commit = git_commit(translation_unit.target_code_root)  # +8
    return new_state, CodeTranslationAgentTranslateClass(2, 2), total_score, commit  # +12


class CodeTranslationAgentTranslateClass:  # +3
    def __init__(self, beam_width, branching):  # +9
        self.beam_width = beam_width  # +5
        self.branching = branching  # +5

        self.called = False  # +5

    def __call__(self, cur_state, cur_commit, cur_score):  # +11
        # Get used variables from `cur_state`
        target_code_root = cur_state["target_code_root"]  # +6
        translation_unit = cur_state["translation_unit"]  # +6
        total_score = cur_state["total_score"]  # +6
        generate_stubs_success = cur_state["generate_stubs_success"]  # +6

        if not self.called:  # +6
            checkout_new_branch(cur_commit)  # +4

            self.translate_class_results = translate_class(translation_unit, beam_width=self.beam_width, default_branching=self.branching)  # x (-0+20)
            self.output_idx = 0  # +5
            self.called = True  # +5

        (pass_tests_count, translate_success_count, num_methods_to_translate, new_branch), _ = self.translate_class_results[self.output_idx]  # x (see above)

        # "+1" to prevent agent from "cheating" (have very few e.g. zero stubs to implement)
        total_score += pass_tests_count / (num_methods_to_translate + 1)  # +9

        ...  # Log results

        # Increment result_idx
        self.output_idx += 1  # +5

        # Store new variables to `new_state`
        new_state = cur_state.copy()  # +6
        new_state["total_score"] = total_score  # +6
        new_state["results"] = results  # +6
        # for translation_unit in translation_units:  # -5
        new_state["translation_unit_idx"] += 1  # +6

        # git commit; decide next step
        if new_state["translation_unit_idx"] == len(new_state["translation_units"]):  # +12
            commit = None  # +3
            next_step = code_translation_agent_postlude  # +3
        else:  # +2
            new_translation_unit = new_state["translation_units"][new_state["translation_unit_idx"]]  # +10
            commit = git_commit(  # +4
                translation_unit.target_code_root,  # +4
                f"Begin {new_translation_unit.source_class_path} translation.",  # +10
            )  # +1
            next_step = code_translation_agent_generate_stubs  # +3
        return new_state, next_step, total_score, commit  # +8


def code_translation_agent_postlude(cur_state, cur_commit, cur_score):  # +9
    # Use beam search to translate a repository
    target_code_root = cur_state["target_code_root"]  # +6
    repo = cur_state["repo"]  # +6

    ...  # Final logging and saving

    return_value = final_commit  # x (see below)
    return return_value, None, cur_score, None  # x (-0+8)


def code_translation_agent(source_code_root, target_code_root, args, beam_width, default_branching):  # x (-0+4)
    # Use beam search to translate a class method-by-method

    init_state = {  # +3
        "source_code_root": source_code_root,  # +5
        "target_code_root": target_code_root,  # +5
        "args": args,  # +5
    }  # +1
    init_step = code_translation_agent_prelude_1  # +3
    init_commit = None  # +3
    init_score = 0.0  # +5

    beam = [init_step(init_state, init_commit, init_score)]  # +11
    results = []  # +3
    while len(beam) > 0:  # +8
        new_program_states_list = []  # +3
        for state, step, score, commit in beam:  # +11
            branching = default_branching if not isinstance(step, CodeTranslationAgentTranslateClass) else step.branching * step.beam_width  # +19
            new_program_states = [step(state, commit, score) for _ in range(branching)]  # +18
            new_program_states.sort(key=lambda x: x[2], reverse=True)  # +17
            new_program_states_list.append(new_program_states)  # +6
        not_done_new_program_states = []  # +3
        for i in range(len(new_program_states_list[0])):  # +11
            # random permutation of indices to break ties
            for j in np.random.permutation(len(new_program_states_list)):  # +13
                new_program_state = new_program_states_list[j][i]  # +8
                new_state, new_step, new_score, new_commit = new_program_state  # +9
                if new_step is None:  # +5
                    results.append((new_state, new_score))  # +8
                else:  # +2
                    not_done_new_program_states.append(new_program_state)  # +6
        not_done_new_program_states.sort(  # +4
            key=lambda program_state: program_state.score, reverse=True  # +12
        )  # +1
        beam = not_done_new_program_states[:beam_width]  # +6
    return max(results, key=lambda x: x[2])[0]  # +16


code_translation_agent(..., beam_width=3, default_branching=3)  # x (-0+8)
\end{lstlisting}
\newpage

\subsection{Case Study 2: Hypothesis Search Agent} \label{app:python_case1}

\textit{Base agent:}

\begin{lstlisting}[basicstyle=\ttfamily\small, caption=Simple 2-step agent for ARC (base), label=lst:hypothesis_base]
def two_step_agent(task_info):
    # Step 1: Get natural language hypothesis
    ...
    hypothesis = hypothesis_agent([task_info], hypothesis_instruction)

    # Step 2: Implement the hypothesis in code
    ...
    code = solver_agent([task_info, hypothesis], solver_instruction)
    return get_test_output(code)


two_step_agent(task_info)
\end{lstlisting}

\newpage

\textit{With \oursimpl{}:}

\begin{lstlisting}[basicstyle=\ttfamily\small, caption={Parallelized BFS in \oursimpl{}, 2 branchpoints}, label=lst:hypothesis_ours]
import encompass  # +2


@encompass.compile  # +4
def two_step_agent(task_info):
    branchpoint()  # +2
    # Step 1: Get natural language hypothesis
    ...
    hypothesis = hypothesis_agent([task_info], hypothesis_instruction)

    branchpoint()  # +2
    # Step 2: Implement the hypothesis in code
    ...
    code = solver_agent([task_info, hypothesis], solver_instruction)

    # Evaluate
    percent_correct = run_validation(code)  # +6
    record_score(percent_correct)  # +4
    if percent_correct == 1:  # +5
        early_stop_search()  # +2

    return get_test_output(code)


two_step_agent(task_info).search("parallel_bfs", default_branching=8)  # x (-0+9)
\end{lstlisting}

\newpage

\textit{Without \oursimpl{}:}
The code devoted to parallelization obscures the underlying agent logic.

\begin{lstlisting}[basicstyle=\ttfamily\small, caption={Parallelized BFS implemented in plain Python}, label=lst:hypothesis_parallel]
from concurrent.futures import ThreadPoolExecutor, as_completed  # +8


def two_step_agent(task_info, branching):  # x (-0+2)
    results = []  # +3
    full_solved = False  # +3

    with ThreadPoolExecutor() as executor:  # +6

        def run_one_forward_pass():  # +3
            if full_solved:  # +3
                return  # +1
            # Step 1: Get natural language hypothesis
            ...  # ->2
            hypothesis = hypothesis_agent([task_info], hypothesis_instruction)  # ->2

            def implement_in_code():  # +3
                nonlocal full_solved  # +2

                if full_solved:  # +3
                    return  # +1

                # Step 2: Implement the hypothesis in code
                ...  # ->3
                code = solver_agent([task_info, hypothesis], solver_instruction)  # ->3

                # Evaluate
                percent_correct = run_validation(code)  # +6
                if percent_correct == 1:  # +5
                    full_solved = True  # +3
                results.append((get_test_output(code), percent_correct))  # x (-1+7)

            futures = [executor.submit(implement_in_code) for _ in range(branching)]  # +16
            for future in as_completed(futures):  # +7
                future.result()  # +4

        futures = [executor.submit(run_one_forward_pass) for _ in range(branching)]  # +16
        for future in as_completed(futures):  # +7
            future.result()  # +4

    return max(results, key=lambda x: x[1])[0]  # +16


two_step_agent(task_info, branching=8)  # x (-0+4)
\end{lstlisting}

\newpage

\subsection{Case Study 3: Reflexion Agent} \label{app:python_case2}

\textit{Base agent:}

\begin{lstlisting}[basicstyle=\ttfamily\small, caption=Reflexion agent (base), label=lst:reflexion_base]
def reflexion_agent(task_info, internal_tests, max_iters):
    # first attempt
    code = solver_agent(task_info)
    percent_correct, feedback = run_validation(code, internal_tests)

    # if solved, exit early
    if percent_correct == 1.0:
        return code

    for cur_iter in range(1, max_iters):
        # self-reflect and apply to next attempt
        reflection = self_reflection_agent(code, feedback)
        code = solver_agent(task_info, code, feedback, reflection)
        percent_correct, feedback = run_validation(code, internal_tests)

        # if solved, exit early
        if percent_correct == 1.0:
            return code

    return code


reflexion_agent(...)
\end{lstlisting}

\newpage

\textit{With \oursimpl{}:}

\begin{lstlisting}[basicstyle=\ttfamily\small, caption={Reexpand best-first search in \oursimpl{}, 2 branchpoints}, label=lst:reflexion_ours]
import encompass  # +2


@encompass.compile  # +4
def reflexion_agent(task_info, internal_tests, max_iters):
    record_score(0.2)  # +6
    branchpoint()  # +2
    # first attempt
    code = solver_agent(task_info)
    percent_correct, feedback = run_validation(code, internal_tests)
    record_score(percent_correct)  # +4
    optional_return(code)  # +4

    # if solved, exit early
    if percent_correct == 1.0:
        early_stop_search()  # x (-2+2)

    for cur_iter in range(1, max_iters):
        branchpoint()  # +2
        # self-reflect and apply to next attempt
        reflection = self_reflection_agent(code, feedback)
        code = solver_agent(task_info, code, feedback, reflection)
        percent_correct, feedback = run_validation(code, internal_tests)
        record_score(percent_correct)  # +4
        optional_return(code)  # +4

        # if solved, exit early
        if percent_correct == 1.0:
            early_stop_search()  # x (-2+2)

    return code


reflexion_agent(...).search("reexpand_best_first", max_num_results=5)  # x (-0+9)
\end{lstlisting}

\newpage

\textit{Without \oursimpl{}:}
Defining separate actions for search obscures the ordering of actions.

\begin{lstlisting}[basicstyle=\ttfamily\small, caption={Reexpand best-first search implemented in plain Python}, label=lst:reflexion_plain]
from queue import PriorityQueue  # +4


def get_initial_attempt(task_info, internal_tests, max_iters):  # +9
    # first attempt
    code = solver_agent(task_info)
    percent_correct, feedback = run_validation(code, internal_tests)

    # if solved, exit early
    if percent_correct == 1.0:
        early_stop = True  # x (-2+3)

    next_step = do_one_reflexion  # +3
    return next_step, early_stop, percent_correct, code, feedback, 1  # +12


def do_one_reflexion(task_info, internal_tests, max_iters, code, feedback, cur_idx):  # +15
    # self-reflect and apply to next attempt
    reflection = self_reflection_agent(code, feedback)  # <-1
    code = solver_agent(task_info, code, feedback, reflection)  # <-1
    percent_correct, feedback = run_validation(code, internal_tests)  # <-1

    # if solved, exit early
    if percent_correct == 1.0:  # <-1
        early_stop = True  # <-1  # x (-2+3)

    next_idx = cur_idx + 1  # x (-8+4)
    next_step = None if next_idx == max_iters else do_one_reflexion  # +9
    return next_step, early_stop, percent_correct, code, feedback, next_idx  # +12


# Apply best-first search choosing the highest-scoring state
# to apply an action
def reflexion_agent(task_info, internal_tests, max_iters, max_num_results):  # x (-0+2)
    init_program_state = ()  # +3
    init_step = get_initial_attempt  # +3
    program_states_to_expand = PriorityQueue()  # +4
    program_states_to_expand.put((init_step, init_program_state))  # +8
    percent_correct = None  # +3
    finished = False  # +3
    num_results = 0  # +3
    results = []  # +3
    while not program_states_to_expand.empty() and not finished:  # +10
        step, program_state = program_states_to_expand.pop()  # +8
        program_states_to_expand.put(program_state)  # put it back  # +6
        next_step, early_stop, percent_correct, code, feedback, next_idx = step(task_info, internal_tests, max_iters, *program_state)  # +23
        results.append((code, percent_correct))  # +8
        if early_stop:  # +3
            break  # +1
        if next_step is not None:  # +6
            program_states_to_expand.put((next_step, (code, feedback, next_idx)))  # +13
        num_results += 1  # +3
        if num_results >= max_num_results:  # +5
            break  # +1
    return max(results, key=lambda x: x[1])[0]  # x (-1+15)


reflexion_agent(..., max_num_results=5)  # x (-0+4)
\end{lstlisting}

\newpage

\end{document}